\documentclass{article}
\usepackage[accepted]{icml2023}
\usepackage{adjustbox}
\usepackage{wrapfig}
\usepackage{multicol}
\usepackage{microtype}
\usepackage{graphicx}
\usepackage{subfigure}
\usepackage{booktabs}
\usepackage{hyperref}
\usepackage{algorithm}
\usepackage{algpseudocode}
\usepackage{amsmath}
\usepackage{amssymb}
\usepackage{mathtools}
\usepackage{amsthm}
\usepackage{nicematrix}
\usepackage[capitalize,noabbrev]{cleveref}
\usepackage[inline, shortlabels]{enumitem}
\usepackage{multirow}
\usepackage{amsthm}
\usepackage{amsmath,amsfonts,bm}
\usepackage{nicefrac}
\usepackage{array}

\definecolor{bluex}{rgb}{0.27, 0.42, 0.81}

\theoremstyle{plain}

\theoremstyle{definition}

\theoremstyle{remark}

\DeclareMathOperator{\softmax}{softmax}

\newcommand{\bR}{\mathbb{R}}

\newcommand{\bP}{\mathbb{P}}
\newcommand{\bT}{\mathbb{T}}

\newcommand{\hE}{\mathcal{E}}

\newcommand{\hL}{\mathcal{L}}
\newcommand{\hM}{\mathcal{M}}

\newcommand{\hQ}{\mathcal{Q}}

\newcommand{\hS}{\mathcal{S}}
\newcommand{\hT}{\mathcal{T}}

\newcommand{\hV}{\mathcal{V}}

\newcommand{\hY}{\mathcal{Y}}

\newcommand{\va}{{\bf a}}

\newcommand{\vh}{{\bf h}}

\newcommand{\vk}{{\bf k}}

\newcommand{\vq}{{\bf q}}

\newcommand{\vv}{{\bf v}}

\newcommand{\vx}{{\bf x}}

\newcommand{\tw}{\text{\ttfamily w}}

\newcommand{\vK}{{\bf K}}

\newcommand{\vV}{{\bf V}}

\newcommand{\vTheta}{{\boldsymbol \Theta}}

\newcommand{\vphi}{{\boldsymbol \phi}}

\newcommand{\vtheta}{{\boldsymbol \theta}}

\newcommand*{\tttext}[1]{\text{\ttfamily #1}}

\newcommand*{\Scale}[2][4]{\scalebox{#1}{$#2$}}%

\usepackage{pifont}% http://ctan.org/pkg/pifont
\newcommand{\cmark}{\ding{51}}%
\newcommand{\xmark}{\ding{55}}%

\algrenewcommand\textproc{}

\icmltitlerunning{Effective Structured Prompting by Meta-Learning and Representative Verbalizer}
\begin{document}
	
	\twocolumn[
	\icmltitle{Effective Structured Prompting \\
		by Meta-Learning and Representative Verbalizer}
	\icmlsetsymbol{equal}{*}
	
	\begin{icmlauthorlist}
		\icmlauthor{Weisen Jiang}{sustech,hkust}
		\icmlauthor{Yu Zhang}{sustech,pcl}
		\icmlauthor{James T. Kwok}{hkust}
	\end{icmlauthorlist}
	
	\icmlaffiliation{sustech}{Guangdong Provincial Key Laboratory of Brain-inspired Intelligent Computation, Department of Computer Science and Engineering, Southern University of Science and Technology}
	\icmlaffiliation{hkust}{Department of Computer Science and Engineering, Hong Kong University of Science and Technology}
	\icmlaffiliation{pcl}{Peng Cheng Laboratory}
	
	\icmlcorrespondingauthor{Yu Zhang}{yu.zhang.ust@gmail.com}
	\icmlkeywords{meta-learning, prompting, prompt tuning, prompt learning}
	
	\vskip 0.3in
	]
	
	\printAffiliationsAndNotice{}
	
	%%%%%%%%%%%%%%%%%%%
	%%%%%%%%%%%%%%%%%%%
	%%%%%%%%%%%%%%%%%%%
	%   section: abstract
	%%%%%%%%%%%%%%%%%%%
	%%%%%%%%%%%%%%%%%%%
	%%%%%%%%%%%%%%%%%%%
	\begin{abstract}
		Prompt tuning for pre-trained masked
		language models (MLM)
		has shown promising performance in
		natural language processing tasks with few labeled examples.
		It tunes a \textit{prompt} for the downstream task,
		and a \textit{verbalizer} is used to bridge the predicted token and label prediction.
		Due to the limited training data,
		prompt initialization
		is crucial for prompt tuning.
		Recently, MetaPrompting~\citep{hou2022metaprompting}
		uses 
		meta-learning to learn a shared initialization 
		for all task-specific prompts.
		However,
		a single initialization
		is insufficient to obtain good prompts for all tasks
		and samples when the tasks are complex.	
		Moreover,
		MetaPrompting requires tuning the whole MLM,
		causing a heavy burden on computation and memory as the MLM is usually large.
		To address these issues,
		we use a prompt pool to extract more task knowledge 
		and construct instance-dependent prompts via attention.
		We further propose a
		novel soft verbalizer (RepVerb)
		which constructs label embedding from feature embeddings directly.
		Combining meta-learning the prompt pool and RepVerb,
		we propose MetaPrompter
		for effective structured prompting.
		MetaPrompter
		is parameter-efficient as only the pool is required to be tuned.
		Experimental results demonstrate that 
		MetaPrompter performs better than the recent state-of-the-arts
		and RepVerb outperforms existing soft verbalizers.
		Code: \url{https://github.com/ws-jiang/MetaPrompter_public}
	\end{abstract}
	
	%%%%%%%%%%%%%%%%%%%
	%%%%%%%%%%%%%%%%%%%
	%%%%%%%%%%%%%%%%%%%
	%   section: intro
	%%%%%%%%%%%%%%%%%%%
	%%%%%%%%%%%%%%%%%%%
	%%%%%%%%%%%%%%%%%%%
	\section{Introduction}
	\label{sec:intro}
	
	In recent years, large pre-trained language models
	have achieved great success in solving a variety of downstream
	tasks 
	\citep{howard2018universal, devlin2019bert, yang2019xlnet,conneau2019cross, song2020mpnet, guo2020jointly,raffel2020exploring, brown2020language,  lester2021power,cui2022prototypical}.
	Though fine-tuning
	the whole model~\citep{howard2018universal,
		devlin2019bert}
	is effective and widely-used, 
	optimizing and storing all the task-specific parameters can be 
	compute- and memory-expensive
	when the model is large
	(e.g., GPT-3~\citep{brown2020language}
	contains $100+$ billion parameters).
	To alleviate this issue,
	many approaches have been proposed. Examples include 
	adapter tuning~\citep{houlsby2019parameter, lin2020exploring, hu2021lora} and
	prompt learning~\citep{radford2019language, shin2020autoprompt,brown2020language, lester2021power, liu2021gpt, li2021prefix, 
		liu2022p, prasad2022grips, liu2022makes}.
	However, prompt learning is more preferable due to its effectiveness and also that it
	can be easily plugged into a pre-trained MLM
	without invasive modification~\citep{li2021prefix, hambardzumyan2021warp, he2022hyperprompt, sun2022bbtv2}.
	
	\textit{Prompt learning}
	formulates the downstream task
	as a cloze-style MLM problem.
	It is useful
	for few-shot tasks due to its effectiveness,
	parameter-efficiency,
	and plug-and-play
	nature~\citep{radford2019language,
		brown2020language,  liu2022makes}.
	Specifically,
	prompt learning wraps
	an input text with a 
	discrete
	\textit{prompt} (e.g., ``\tttext{Topic is [MASK]}'')
	and feeds it to the MLM to predict a token at the \tttext{[MASK]} position.
	A \textit{verbalizer}~\citep{lester2021power,ding2021openprompt, hu2022knowledgeable} 
	then maps the predicted token to the label.
	However,	designing an
	effective prompt requires
	a good understanding of the downstream tasks.
	
	Recently,
	\textit{prompt tuning}~\citep{lester2021power, liu2021gpt, zhang2022differentiable}
	proposes to
	wrap the input embedding with a
	\textit{continuous} prompt.
	To reduce the number of parameters to be learned,
	the MLM 
	is kept 
	frozen.
	The continuous prompt can be further combined 
	with discrete tokens
	to form a \textit{template}
	\citep{liu2021gpt,schick2021exploiting, ding2021openprompt}.
	
\begin{figure}[!t]
		\centering
		\includegraphics[width=0.48\textwidth]{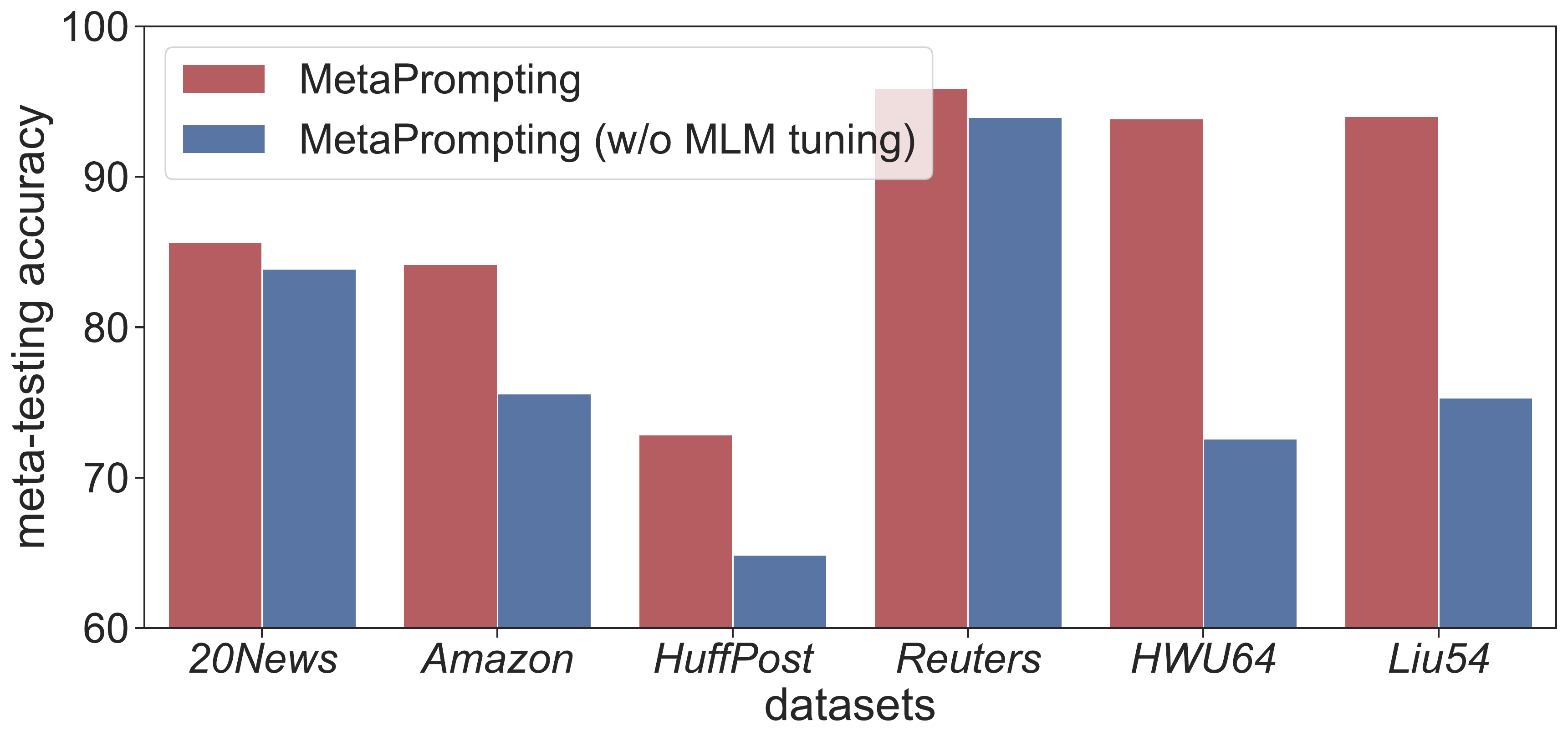}
		\vskip -.2in
		\caption{5-way 5-shot classification meta-testing accuracy of MetaPrompting with or without MLM tuning
			on six data sets.}
		\label{fig:metaprompting-wo-tuning-LM}
		\vskip -.2in
	\end{figure}

	Prompt tuning can be sensitive to initialization \citep{lester2021power}.
	Recently, a
	number of approaches have been proposed to alleviate this problem
	\cite{lester2021power, li2022learning,vu2022spot}. 
	In particular,
	MetaPrompting~\citep{hou2022metaprompting}
	is the state-of-the-art
	that uses
	\textit{meta-learning}~\citep{bengio1991, thrun1998learning, Finn2017}
	to learn a meta-initialization for all task-specific
	prompts.
	However, MetaPrompting 
	suffers from three problems.
	\begin{enumerate*}[(i), series = tobecont, itemjoin = \quad]
		\item When the tasks are complex, it is challenging
		to obtain good prompts for all tasks and samples from a single meta-initialized
		prompt.
		\item MetaPrompting uses
		a hand-crafted verbalizer. However,
		selecting
		good label tokens 
		for the hand-crafted verbalizer
		is labor-intensive and not scalable
		for a large label set.
		\item 
		MetaPrompting requires
		expensive tuning the whole MLM.
		Figure \ref{fig:metaprompting-wo-tuning-LM} shows
		a large gap in
		meta-testing accuracies
		with and without MLM
		tuning
		(experimental details are in Section~\ref{sec:expt}).
	\end{enumerate*}
	
	In this paper,
	we use a pool
	of multiple prompts 
	\citep{li2022learning, wang2022dualprompt, wang2022learning}
	to extract task knowledge from meta-training tasks, and then construct instance-dependent prompts
	as weighted combinations of all the prompts in the pool via attention \citep{vaswani2017attention}.
	The attention's  query vector is the instance's feature embedding.
	The prompt pool is the shared meta-knowledge and learned by the MAML algorithm~\citep{Finn2017}.
	Specifically,
	given a task with a support set and a query set,
	the base learner takes the meta-parameter and the support set to build a task-specific prompt pool,
	then the meta-learner optimizes the meta-parameter on the query set.
	Meta-learning a prompt pool is more flexible than meta-learning only a single
	prompt initialization (as in
	MetaPrompting),
	and allows better adaptation of complex tasks.
	Moreover, as only the prompt pool is tuned, it is much more parameter-efficient 
	than MetaPrompting
	(with $1000 \times $ fewer parameters).
	
	We also propose a novel soft verbalizer called
	{\em representative verbalizer\/} (RepVerb),
	which constructs label embeddings
	by averaging feature embeddings of the corresponding
	training samples.
	Unlike manually-designed verbalizers,
	RepVerb does not incur human effort for
	label token
	selection. 
	Moreover,
	as RepVerb does not require learning any additional parameters,
	empirical results in Section \ref{subsec:expt-verb} demonstrate that 
	RepVerb is more effective than
	the soft verbalizers in
	WARP~\citep{hambardzumyan2021warp}, DART~\citep{zhang2022differentiable}, ProtoVerb~\citep{cui2022prototypical}.
	Besides,
	the feature embedding learned by RepVerb is more discriminative.
	
	The whole procedure, which
	combines meta-learning the structured prompts and RepVerb,
	is called \textbf{MetaPrompter} in the sequel.
	Experiments are performed on six widely used classification data sets.
	Results  demonstrate that RepVerb outperforms existing soft verbalizers, and
	is also beneficial to other prompt-based methods such as MetaPrompting.
	Moreover, MetaPrompter
	achieves better performance than the recent state-of-the-arts.
	
	Our contributions are summarized as follows:
	\begin{enumerate*}[(i), series = tobecont, itemjoin = \quad]
		\item We propose a
		parameter-efficient algorithm MetaPrompter for effective structured prompting.
		\item 
		We propose 
		a simple and effective soft verbalizer (RepVerb).
		\item Experimental results demonstrate the effectiveness and parameter-efficiency of MetaPrompter.
	\end{enumerate*}
	
	%%%%%%%%%%%%%%%%%%%
	%%%%%%%%%%%%%%%%%%%
	%%%%%%%%%%%%%%%%%%%
	%   section: related work
	%%%%%%%%%%%%%%%%%%%
	%%%%%%%%%%%%%%%%%%%
	%%%%%%%%%%%%%%%%%%%
	
	\section{Preliminaries and Related Work}
	\label{sec:related-work}
	
	\subsection{Prompt Learning}
	\label{sec:2a}
	
	\vskip -.05in
	Recently,  it is common to use a
	pre-trained MLM $\hM(\cdot; \vphi)$,
	with parameter $\vphi$, for various downstream tasks such as
	language understanding~\citep{dong2019unified,yang2019xlnet, song2020mpnet},
	machine translation~\citep{conneau2019cross, guo2020jointly},
	and
	text classification~\citep{brown2020language, lester2021power, liu2022p}.
	Given a raw sentence represented as a sequence of $n$ tokens $(x_1, \dots, x_n)$,
	the MLM takes $\vx=(\tttext{[CLS]}, x_1, \dots, x_n, \tttext{[SEP]})$  as input
	(where \tttext{[CLS]}
	is the start token
	and \tttext{[SEP]}
	is the separator), and encodes  it into a sequence of hidden
	representations $(\vh_{\tttext{[CLS]}}, \vh_1, \dots, \vh_n,
	\vh_{\tttext{[SEP]}})$.  In standard fine-tuning~\citep{howard2018universal,
		devlin2019bert}, an extra classifier (e.g., a fully connected layer with softmax
	normalization) is added on top of $\vh_{\tttext{[CLS]}}$ to predict the label
	distribution.
	This classifier,
	together with $\vphi$,
	are tuned to maximize the probability of correct
	labels.  As language models are large (e.g., $175$ billion parameters in
	GPT-3~\citep{brown2020language}), fine-tuning all parameters can cause a heavy burden on
	computation and memory.
	
	On the other hand,
	prompt learning~\citep{brown2020language, shin2020autoprompt, ding2021openprompt}
	freezes the
	pre-trained model and formulates the downstream task as a cloze-style MLM problem.
	For example,
	in topic classification,
	``\tttext{Topic is [MASK]}''
	can be used
	as the
	prompt,
	where \tttext{[MASK]} is a special token for prediction.
	The
	\textit{discrete}
	tokens ``\tttext{Topic is}'' are also called
	anchor tokens.
	An
	input text
	$\vx$
	is wrapped with the prompt and mapped to  an input embedding sequence
	$(\hE(\vx), \hE(\tttext{Topic}), \hE(\tttext{is}), \hE(\tttext{[MASK]}))$,
	where $\hE(\cdot)$ denotes the input embedding.
	Designing a suitable prompt requires domain expertise and a good
	understanding of the downstream tasks~\citep{brown2020language,
		sanh2022multitask}. Thus, manually-designed prompts are likely to be sub-optimal.
	
	Unlike discrete prompts, prompt tuning~\citep{lester2021power, liu2021gpt} uses a
	\textit{continuous} prompt $\vtheta \in \bR^{L_p\times d_i}$ (of length $L_p$) to
	directly wrap the input embedding sequence as $(\hE(\vx), \vtheta, \hE(\tttext{[MASK]}))$.
	This can be further combined with anchor tokens
	to form a \textit{template}
	\citep{liu2021gpt, schick2021exploiting, ding2021openprompt}:
	\begin{equation*}
		\tilde{\vx}\equiv \bT(\vx;\vtheta)
		\!\!=\!\! (\hE(\vx), \vtheta, \hE(\tttext{Topic}),
		\hE(\tttext{is}), \hE(\tttext{[MASK]})).
	\end{equation*}  
	The MLM then outputs
	the hidden embedding
	$\vh_{\tttext{[MASK]}}(\tilde{\vx}) \in \bR^{d_o}$
	of \tttext{[MASK]},
	and infers the token to be filled at the
	\tttext{[MASK]} position.
	
	A \textit{verbalizer}~\citep{lester2021power,ding2021openprompt, hu2022knowledgeable}
	bridges the prediction at the \tttext{[MASK]}
	position
	and labels
	in prompt learning.
	Specifically,
	it is a \textit{hard} mapping from each label $y$ to a set of label-relevant tokens $\hV_y$. For example,
	for $y=\tttext{SPORTS}$, we can have
	$\hV_y=\{\tttext{sports, football, basketball}\}$.
	Prompt tuning then optimizes\footnote{$\vphi$ can be fixed for parameter-efficiency in prompt learning.}
	$(\vphi, \vtheta)$
	by maximizing the label probability:
	\begin{equation} \label{eq:hard-pred}
		\hat{\bP}(y| \vx; \vphi, \vtheta) \!=\!  \frac{1}{|\hV_y|}  \!\! \sum_{\tw\in \hV_y} \!\! \bP_\hM(\tttext{[MASK]} = \tw | \bT(\vx; \vtheta)),
	\end{equation} 
	where $\bP_\hM(\tttext{[MASK]} | \bT(\vx; \vtheta))$ is the probability distribution
	over vocabulary as
	predicted by the MLM
	at the \tttext{[MASK]} position.
	
	The verbalizer
	is crucial to the performance of prompt
	learning~\citep{lester2021power,ding2021openprompt}. However,
	selecting
	label-relevant
	tokens
	requires
	intensive human labor.  To address this problem, search-based
	methods~\citep{schick2020automatically, shin2020autoprompt, gao2021making} try
	to find label tokens
	automatically
	from the training data.
	However, searching in a \textit{discrete} space is  computationally
	intensive~\citep{schick2020automatically, shin2020autoprompt,
		gao2021making}, especially with a
	large number
	of labels or vocabulary.
	Some recent works
	\citep{hambardzumyan2021warp,
		zhang2022differentiable, cui2022prototypical}
	propose \textit{soft} verbalizers,
	which
	map each label to a \textit{continuous}
	embedding
	and predict the label distribution based on the similarities between feature embedding and label embeddings.
	WARP~\citep{hambardzumyan2021warp} and DART~\citep{zhang2022differentiable} obtain
	this
	label embedding by supervised learning,
	while
	ProtoVerb~\citep{cui2022prototypical}
	uses contrastive learning~\citep{chen2020simple,
		tian2020makes}.
	However,
	learning the embedding
	$\vv_y \in \bR^{d_o}$ for each label $y$
	can be challenging
	in the few-shot learning  setting
	\citep{gao2019hybrid, bao2020few, han2021meta, chen2022contrastnet, hou2022metaprompting},
	as the number of samples per class is typically much smaller than $d_o$ (e.g., $d_o=768$ for BERT~\citep{devlin2019bert}).
	
	\subsection{Meta-Learning for Prompt Learning}
	
	In meta-learning~\citep{bengio1991, thrun1998learning},
	a collection $\hT$ of tasks are used to learn a shared
	meta-parameter.  
	Each task $\tau\in \hT$ has a support set $\hS_{\tau}$ and a
	query set $\hQ_\tau$. Let $\hY_\tau$ be the label set of $\tau$.
	Typical meta-learning algorithms can be
	metric-based \citep{ vinyals2016matching, snell2017prototypical,
		bertinetto2018meta, lee2019meta}, memory-based
	\citep{santoro2016meta, munkhdalai2017meta}, or
	optimization-based~\citep{Finn2017, rajeswaran2019meta, raghu2020rapid,ye2021multiobjective,
		jiang2021effective, jiang2022subspace, Flennerhag2022}.
	In general, the optimization-based approach is preferred due to its simplicity and
	effectiveness.
	A representative algorithm
	is
	model-agnostic meta-learning (MAML)~\citep{Finn2017}.
	
	As prompt tuning is sensitive to prompt initialization in few-shot tasks~\citep{lester2021power}, meta-learning can be used to  search for a good initialization.
	MetaPrompting~\citep{hou2022metaprompting}
	uses
	MAML
	to learn a meta-initialization for the task-specific
	prompts.
	At iteration $t$, the base learner takes
	a task $\tau$ and meta-parameter $(\vphi_{t-1}, \vtheta_{t-1})$, and builds a
	task-specific model
	$(\vphi_{t, J}, \vtheta_{t, J})$ by
	performing
	$J$ gradient updates on the support set with
	step size $\alpha$ and initialization $(\vphi_{t, 0}, \vtheta_{t, 0}) \equiv (\vphi_{t-1}, \vtheta_{t-1})$:
	\begin{align*}
		&(\vphi_{t, j}, \vtheta_{t, j}) = (\vphi_{t, j-1}, \vtheta_{\vx,j-1}) \nonumber \\
		& \; +\!  \alpha \nabla_{(\vphi_{t, j-1}, \vtheta_{\vx,j-1})}\!\! \!\!\!\!\! \sum_{(\vx, y) \in \hS_\tau} \!\! \!\! \!\!\log  \hat{\bP}(y| \vx; \vphi_{t, j-1}, \vtheta_{\vx,j-1}).
	\end{align*} 
	The meta-learner then updates the meta-initialization by
	maximizing the log-likelihood objective
	on the query set with
	step size $\eta$:
	\begin{align*}
		& (\vphi_{t}, \vtheta_{t}) = (\vphi_{t-1}, \vtheta_{t-1}) \\
		&\;\;\;\;\; + \eta \nabla_{(\vphi_{t-1}, \vtheta_{t-1})} \sum_{(\vx, y) \in \hQ_\tau}\log \hat{\bP}(y| \vx; \vphi_{t, J}, \vtheta_{t, J}).
	\end{align*}
	Though MetaPrompting achieves state-of-the-art performance in the few-shot
	classification experiments
	\citep{hou2022metaprompting},
	it suffers from the three problems discussed in Section \ref{sec:intro}. 
	\begin{enumerate*}[(i), series = tobecont, itemjoin = \quad]
		\item When the tasks are complex, 
		it is challenging to use
		a single meta-initialized prompt for adaptation to the various tasks.
		\item MetaPrompting uses
		a hand-crafted verbalizer,
		which is labor-intensive and not scalable as discussed in Section~\ref{sec:2a}.
		\item MetaPrompting needs to tune the MLM parameters, and thus is not parameter-efficient.
	\end{enumerate*}
	
	%%%%%%%%%%%%%%%%%%%
	%%%%%%%%%%%%%%%%%%%
	%%%%%%%%%%%%%%%%%%%
	%   section: method
	%%%%%%%%%%%%%%%%%%%
	%%%%%%%%%%%%%%%%%%%
	%%%%%%%%%%%%%%%%%%%
	
	\section{Proposed Method}
	\label{sec:method}
	
	\begin{algorithm*}[!h]
		\caption{\underline{Rep}resentative \underline{Verb}alizer (RepVerb).}
		\label{alg:repverb}	
		\begin{multicols}{2}
			\begin{algorithmic}[1]
				% proc 1
				\Procedure{\textsf{ComputeLabelEmbedding}}{$\hS_{\tau}$}:
				\State compute $\vh_{\tttext{[MASK]}}(\tilde{\vx})$
				for $(\vx,\cdot) \in \hS_{\tau}$;
				\State compute $\vv_y$ by
				\eqref{eq:vy} for $y \in \hY_\tau$;
				\EndProcedure
				\setcounter{ALG@line}{0}
				% proc 2
				\Procedure{\textsf{Predict}}{$\vx;  \vv_y: y\in \hY_\tau$}
				\State compute $\vh_{\tttext{[MASK]}}(\tilde{\vx})$ for $\vx$; 
				\State
				compute  
				$\tilde{\bP}(y|\vx; \vphi, \vtheta)$ by \eqref{eq:soft-pred};
				\EndProcedure	
			\end{algorithmic}
		\end{multicols}
		\vskip -.1in
	\end{algorithm*}
	
	In 
	Section \ref{sec:repverb},
	we first propose a
	novel and effective soft verbalizer (representative verbalizer)
	without inducing additional parameters.
	Moreover, while MetaPrompting uses a single prompt initialization to build
	task-specific prompts,  we propose
	in Section~\ref{sec:MetaPrompter}
	the extraction of task knowledge into a pool of multiple prompts,
	and constructs instance-dependent prompts by attention \citep{vaswani2017attention}.
	
	\begin{algorithm*}[!h]
		\caption{
		%\underline{Meta} Structured-\underline{Prompt}ing with R\underline{e}presentative Ve\underline{r}balizer (
		MetaPrompter.}
		\label{alg}
		
		\begin{algorithmic}[1]
			\Require prompt length $L_p$; size of prompt pool $K$; $\lambda=0.5$;
			step sizes $\alpha, \eta$;
			meta-parameters $(\vK, \vTheta)$;
			query function $q(\cdot)$;
			\For{ $t=1, \dots, T$}
			\State sample a task $\tau = (\hS_{\tau}, \hQ_\tau) \in \hT$;
			%%%%%%%%%%%%%%%%%%
			%%%%%%%%%%%%%%%%%%
			%%%%%%%%%%%%%%%%%%
			% base-learner
			\State \underline{\textit{base learner:}}
			\State $(\vK_{t, 0}, \vTheta_{t,0}) \equiv ( \vK_{t-1}, \vTheta_{t-1})$;
			\label{step:init}
			\For{$j=1,\dots, J$}
			\For{$(\vx, y) \in \hS_{\tau}$}
			\State compute $\vq_\vx$ by $q(\cdot)$;
			\label{step:instance-prompt-1}
			\State $\vtheta_{\vx,j}(\vK_{t, j-1}, \vTheta_{t, j-1})=  \softmax (\vK_{t,j-1}\vq_\vx)^\top \vTheta_{t,j-1} $;
			\label{step:instance-prompt-2}
			\State feed $\tilde{\vx} \equiv \bT(\vx; \vtheta_{\vx,j})$ into $\hM$, obtain $\vh_{\tttext{[MASK]}}(\tilde{\vx})$, and
			$\hat{\bP}(y| \vx;\vtheta_{\vx,j})$ by \eqref{eq:hard-pred};
			\label{step:hard-verb}
			\EndFor
			\State call \textproc{\textsf{ComputeLabelEmbedding}}($\hS_\tau$) of Algorithm \ref{alg:repverb} to obtain $\{\vv_y: y \in \hY_\tau\}$;
			\label{step:soft-verb-1}
			\State for $\!(\vx,\! y) \!\in\! \hS_{\tau}$, 
			call  \textproc{\textsf{Predict}}$(\vx; \! \vv_y \!:\! y\!\in\! \hY_\tau\!)$ of Algorithm \ref{alg:repverb}
			to obtain $\tilde{\bP}(y| \vx; \vtheta_{\vx,j}\!)$,
			and compute $\bP(y|\vx;\!\vtheta_{\vx,j}\!)$
			by  \eqref{eq:total-pred};\!\!\!\!
			\label{step:soft-verb-2}
			\State   $	\hL(\hS_{\tau}; \vK_{t,j-1},\vTheta_{t, j-1})=  -\sum_{(\vx, y) \in \hS_\tau} \!\!\log \bP(y|\vx; \!\vtheta_{\vx,j})$;
			\label{step:support-loss}
			\State   $( \vK_{t,j},\vTheta_{t,j}) = (  \vK_{t,j-1},\vTheta_{t,j-1}) - \alpha \nabla_{ (\vK_{t,j-1}, \vTheta_{t,j-1})} \hL(\hS_{\tau};  \vK_{t,j-1}, \vTheta_{t,j-1})$;
			\label{step:base update}
			\EndFor \label{alg:base-b}
			%			\State $( \vK_{t,\tau},\vTheta_{t,\tau}) \equiv (\vK_{t,J},\vTheta_{t,J})$; 
			%%%%%%%%%%%%%%%%%%
			%%%%%%%%%%%%%%%%%%
			%%%%%%%%%%%%%%%%%%
			% meta-learner
			\State \underline{\textit{meta-learner:}}
			\For{$(\vx, y) \in \hQ_\tau$}
			\label{alg:meta-a}
			\State compute $\vq_\vx$ by $q(\cdot)$;
			\label{step:meta-instance-prompt-1}
			\State  $\vtheta_{\vx,J}(\vK_{t,J}, \vTheta_{t,J}) = \softmax  ( \vK_{t,J}\vq_\vx)^\top \vTheta_{t,J}$;
			\label{step:meta-instance-prompt-2}
			\State call  \textproc{\textsf{Predict}}$(\vx; \vv_y : y\in \hY_\tau)$ of Algorithm \ref{alg:repverb} to obtain $\tilde{\bP}(y|\vx; \vtheta_{\vx,J}\!)$;
			\label{step:meta-soft-pred}
			\State compute $\hat{\bP}(y|\vx;\ \!\vtheta_{\vx,J}\!)$
			and
			$\bP(y|\vx;  \vtheta_{\vx,J}\!)$ by \eqref{eq:hard-pred} and \eqref{eq:total-pred}, respectively;
			\label{step:meta-pred}
			\EndFor
			\State
			$ \hL(\hQ_{\tau}; \vK_{t,J},\vTheta_{t,J})=-\sum_{(\vx, y) \in \hQ_\tau} \log \bP(y|\vx; \vtheta_{\vx,J})$;
			\label{step:meta-loss}
			\State  $( \vK_{t}, \vTheta_{t}) = (  \vK_{t-1}, \vTheta_{t-1}) - \eta \nabla_{( \vK_{t-1}, \vTheta_{t-1})} \hL(\hQ_{\tau}; \vK_{t,J},\vTheta_{t,J})$;
			\label{alg:meta-update}
			\EndFor \\
			\Return $(\vK_T,\vTheta_{T})$.
		\end{algorithmic}
	\end{algorithm*}
	
	\subsection{Representative Verbalizer
		(RepVerb)}
	\label{sec:repverb}
	
	Instead of explicitly learning an embedding
	$\vv_y$
	for each label $y$
	\citep{hambardzumyan2021warp, cui2022prototypical, zhang2022differentiable}, we propose
	the {\em Representative Verbalizer} (RepVerb),
	which
	constructs
	$\vv_y$
	from
	feature
	embeddings of the corresponding training samples
	(Algorithm~\ref{alg:repverb}).
	It does not require
	learning additional parameters, and
	is thus more effective on limited data as in few-shot learning.
	
	Specifically, let $\hS_{\tau, y}$ be the subset  of samples
	in $\hS_\tau$
	with label $y$.
	For an input $\vx$, we wrap it with the template and feed
	$\tilde{\vx} \equiv \bT(\vx;\vtheta)$ to the pre-trained MLM, and then
	obtain \tttext{[MASK]}'s embedding
	$\vh_{\tttext{[MASK]}}(\tilde{\vx})$ as its feature embedding.
	Similar to ProtoNet~\citep{snell2017prototypical},
	we propose to
	construct $\vv_y$
	for each $y$ by averaging the corresponding
	samples' feature embeddings, as:
	\begin{align}
		\vv_y = \frac{1}{|\hS_{\tau, y}|} \sum_{(\vx, y) \in \hS_{\tau, y}} \vh_{\tttext{[MASK]}}(\tilde{\vx}).
		\label{eq:vy}
	\end{align}  
	To predict the label of
	a given $\vx$,
	we measure the cosine similarity\footnote{Dissimilarity measures, such as the Euclidean distance, can also be used.}
	between $\vh_{\tttext{[MASK]}}(\tilde{\vx})$ and each
	$\vv_y$ ($y \in \hY_\tau$):
	\begin{align}
		\!\!\!\!\!\!	\tilde{\bP}(y|\vx; \vphi,  \vtheta)\! =\! \frac{\exp(\rho \cos(\vv_y,\! \vh_{\tttext{[MASK]}}(\tilde{\vx})))}{\sum_{y' \in \hY_\tau} \!\!\! \exp(\rho\cos(\vv_{y'},\! \vh_{\tttext{[MASK]}}(\tilde{\vx})))}, 
		\label{eq:soft-pred}
	\end{align} 
	where $\rho > 0$ is the temperature.
	When  $\rho\to \infty$,
	$\tilde{\bP}(y|\vx; \vphi,  \vtheta)$
	becomes one-hot; whereas
	when $\rho\to 0$,
	$\tilde{\bP}(y|\vx; \vphi,  \vtheta)$
	becomes
	uniform.
	In the experiments, we set $\rho=10$ as in \citet{Oreshkin2018}.
	
	\subsection{Meta Structured-Prompting}
	\label{sec:MetaPrompter}
	
	In the following, we propose the use of MAML and
	attention mechanism~\citep{vaswani2017attention}
	to meta-learn
	a prompt pool.
	While MetaPrompting
	uses
	task-specific prompts~\citep{hou2022metaprompting},
	we propose the construction of instance-specific prompts,
	which allows more flexibility.
	
	\subsubsection{Meta-Learn a Prompt Pool }
	
	While MetaPrompting uses only a single initialization for the prompt,
	we propose to leverage
	a pool of prompts to extract more task knowledge, which is particularly effective when the tasks
	are complex and very different prompts may be
	needed.
	A prompt pool has $K$
	learnable prompts
	$\{(\vk_i, \vtheta_i): i=1,\dots, K \}$, with
	key $\vk_i\in \bR^{d_o}$
	and
	value $\vtheta_i \in \bR^{L_p\times d_i}$
	\citep{li2022learning, wang2022dualprompt, wang2022learning}.
	Note that the 
	size of 
	the prompt pool is negligible 
	compared with that of the MLM. For example, in our experiments, 
	the MLM has $109.52\times 10^6$ parameters,
	while the prompt pool has only
	$55,296$.
	
	The prompt pool can be considered as shared meta-knowledge.
	Given an input
	$\vx$,
	the attention weights
	between $\vx$ and the $K$ prompts are computed as $\va = \softmax (\frac{\vK \vq_\vx}{\sqrt{d_o}})$,
	where $\softmax(\cdot)$ is the softmax function,
	$\vK = [\vk_1^\top; \dots; \vk_K^\top]$, and
	$\vq_\vx \in \bR^{d_o}$ is
	the
	embedding
	of the \tttext{[MASK]}
	output by
	a
	pre-trained and frozen
	MLM with the wrapped input (e.g., $(\vx. \tttext{ Topic is [MASK]})$)
	\citep{wang2022dualprompt, wang2022learning}.
	Such a mapping from $\vx$ to $\vq_\vx$ is called the query function $q(\cdot)$.
	An instance-dependent prompt is then generated by weighted averaging
	over all the
	values 		($\vtheta_i$'s):
	\begin{align}
		\vtheta_\vx(\vK, \vTheta) = \sum_{i=1}^K a_i \vtheta_i, \label{eq:struct-promp}
	\end{align}
	where
	$\vTheta = [\vtheta_1; \dots; \vtheta_K]$.
	While
	\citet{wang2022dualprompt, wang2022learning}
	only select the
	top-$N$ most similar prompts
	from the pool,
	in
	\eqref{eq:struct-promp}
	all the prompts are
	used
	and updated simultaneously.
	
	The proposed procedure for meta-learning the
	prompt pool
	$(\vK, \vTheta)$, which will be called
	MetaPrompter,
	is
	shown in
	Algorithm~\ref{alg}. The MAML algorithm~\citep{Finn2017} is used here, but other meta-learning algorithms (e.g., Reptile \citep{Nichol2018}, BMG~\citep{Flennerhag2022}) can also be used.
	At iteration $t$, the base learner takes $(\vK_{t-1}, \vTheta_{t-1})$ and a task $\tau$ to
	optimize for a task-specific prompt pool 
	by gradient descent
	(steps \ref{step:init}-\ref{alg:base-b}).
	$(\vK_{t-1}, \vTheta_{t-1})$ is used as the initialization (step \ref{step:init}).
	For each inner iteration $j$,
	$(\vK_{t, j-1}, \vTheta_{t, j-1})$ constructs the
	instance-dependent prompts $\vtheta_{\vx,j}(\vK_{t, j-1}, \vTheta_{t, j-1})$ in
	\eqref{eq:struct-promp} (steps
	\ref{step:instance-prompt-1} and \ref{step:instance-prompt-2}).  
	Next, $\vtheta_{\vx,j}$ is used to predict the label
	probability with a combination of the
	hand-crafted
	verbalizer
	(step \ref{step:hard-verb}) and soft verbalizer
	(steps  \ref{step:soft-verb-1} and \ref{step:soft-verb-2}):
	\begin{align}
		\bP(y|\vx; \vtheta_{\vx,j}) \!=\! (1  - \lambda)	\hat{\bP}(y| \vx;\vtheta_{\vx,j} ) +  \lambda \tilde{\bP}(y|\vx; \vtheta_{\vx,j}), \label{eq:total-pred}
	\end{align} 
	where $\lambda \in[0,1]$ (in the experiments,
	we set $\lambda = 0.5$).
	Let $\hL(\hS_{\tau};\vK_{t,j-1}, \vTheta_{\!t, j-1}) =- \sum_{(\vx, y) \in \hS_\tau}
	\log \bP \left(y| \vx; \vtheta_{\vx, j}\right)$ be the loss on $\hS_{\tau}$ (step \ref{step:support-loss}).
	The base learner builds a task-specific prompt pool $(\vK_{t,J},\! \vTheta_{t, J})$ 
	by taking $J$ gradient updates
	($j=1, \dots, J$) at step~\ref{step:base update}:
	\begin{align*}
		\!\!\!(\vK_{t,j\!},\!\! \vTheta_{\!t, j}\!) \!\!
		=\!\! 	(\vK_{t,j-1\!},\!\! \vTheta_{\!t, j-1}\!) \!\!
		-\! \alpha \!\nabla_{\!\!\Scale[0.4]{ (\vK_{t,j-1\!},\! \vTheta_{\!t, j-1}\!)}}\hL\!(\!\hS_{\tau}; \!\vK_{t,j-1\!}, \!\!\vTheta_{\!t, j-1}\!). \!\!\!
	\end{align*}  
	The meta-learner takes $(\vK_{t,J},\! \vTheta_{t, J})$ and $\hQ_{\tau}$
	to update the meta-parameters (steps \ref{alg:meta-a}-\ref{alg:meta-update}).  For
	$(\vx, y) \in \hQ_{\tau}$, we use $(\vK_{t,J},\! \vTheta_{t, J})$ to generate
	its prompt $\vtheta_{\vx, J} (\vK_{t, J}, \vTheta_{t,
		J})$ (steps \ref{step:meta-instance-prompt-1} and \ref{step:meta-instance-prompt-2}),
	which is used for make prediction $\bP \left(y| \vx;
	\vtheta_{\vx, J}\right)$ (steps \ref{step:meta-soft-pred} and \ref{step:meta-pred}).
	Let $\hL(\hQ_{\tau};\vK_{t,J},\!
	\vTheta_{t, J}) = -\sum_{(\vx, y) \in \hQ_\tau}\log \bP \left(y| \vx;
	\vtheta_{\vx, J}\right)$ be the negative log-likelihood loss on $\hQ_{\tau}$ (step \ref{step:meta-loss}).
	The meta-learner updates the meta-parameters by performing one gradient descent step
	on $\hL(\hQ_{\tau};\vK_{t,J},\! \vTheta_{t, J})$
	at step \ref{alg:meta-update}:
	\begin{align*}
		\!\!\!(\vK_{t},\! \vTheta_{t}\!)
		\!=\! 	(\vK_{t-1},\!\vTheta_{t-1}\!)\!
		- \!\eta \nabla_{\!\!\Scale[0.65]{ (\vK_{t-1\!},\! \vTheta_{\!t-1}\!)}} \hL(\hQ_{\tau}; \!\vK_{t, J}, \!\vTheta_{t, J}\!).
	\end{align*} 
	The meta-gradient $\nabla_{(\vK_{t-1}, \vTheta_{t-1})}  \hL(\hQ_{\tau};
	\vK_{t, J}, \vTheta_{t, J}) = \nabla_{(\vK_{t,J}, \vTheta_{t,J})}
	\hL(\hQ_{\tau}; \vK_{t, J}, \vTheta_{t, J}) \nabla_{( \vK_{t-1}, \vTheta_{t-1})}
	(\vK_{t, J}, \vTheta_{t, J})$ requires back-propagating through the entire
	inner optimization path, which is computationally infeasible for large models and $J$ is large.
	To reduce the computational cost,
	we discard the second-order
	derivative and use the first-order approximation
	$\nabla_{(\vK_{t-1}, \vTheta_{t-1})}  \hL(\hQ_{\tau}; \vK_{t, J}, \vTheta_{t, J})\approx\nabla_{(\vK_{t,J}, \vTheta_{t,J}\!)}  \hL(\!\hQ_{\tau}; \!\vK_{ t, J},\! \vTheta_{t, J}\!)$
	(step \ref{alg:meta-update})
	as in \citep{Finn2017, hou2022metaprompting}.
	
	\subsubsection{Meta-Testing}
	
	Given an unseen task $\tau'=(\hS_{\tau'}, \hQ_{\tau'})$,
	the base learner takes
	$\hS_{\tau'}$
	and $(\vK_T,\vTheta_{T})$
	to build a task-specific prompt pool
	$(\vK_{T,J},\vTheta_{T,J})$ as in steps
	\ref{step:init}-\ref{alg:base-b}.
	This pool is then used to construct instance-dependent prompts $\vtheta_{\vx, J}$ for each
	$(\vx,\cdot) \in \hQ_{\tau'}$.
	The MLM receives the wrapped input $\tilde{\vx} \equiv \bT(\vx; \vtheta_{\vx, J})$
	and predicts the label probability by \eqref{eq:total-pred}.
	
	\subsubsection{MetaPrompter is Parameter-Efficient}
	
	As MetaPrompter only tunes $(\vK, \vTheta)$, the total number of meta-parameters is
	$K(d_o + L_p d_i)$
	(where $d_i$ and $d_o$ are the dimensions of the input and feature embeddings, respectively).
	This is much smaller than that of MetaPrompting (which is  equal to $d_\vphi + L_p
	d_i $, where $d_\vphi$ is the size of $\vphi$), as it requires tuning the whole
	MLM.  For example, in the experiments, 
	we use BERT
	(with $d_o=d_i=768$, $d_\vphi = 109\times 10^6$)
	and 
	$K=L_p= 8$
	in MetaPrompter.

	%%%%%%%%%%%%%%%%%%%%%%%%%%%%%
	%   section: experiment
	%%%%%%%%%%%%%%%%%%%%%%%%%%%%%
	\section{Experiments}
	\label{sec:expt}

	\subsection{Setup}	
	Following \citet{chen2022contrastnet}, we
	perform
	few-shot
	classification
	on six popularly used data sets:
	\begin{enumerate*}[(i), itemjoin = \quad]
		\item
		\textit{20News}~\citep{lang1995newsweeder}, which contains informal discourses from news discussion forums of $20$ topics;
		\item \textit{Amazon}~\citep{he2016ups}, which consists of customer reviews  from $24$
		products.  The task is to classify reviews into product categories;
		\item \textit{HuffPost}~\citep{misra2022news}, which contains news headlines of
		$41$ topics published in the HuffPost between 2012 and 2018.
		These headlines are shorter and less grammatical
		than formal sentences,
		thus are more challenging for classification;
		\item \textit{Reuters}~\citep{lewis1997reuters}, which
		is a collection of Reuters newswire articles of
		$31$ topics from $1996$ to $1997$;
		\item \textit{HWU64}~\citep{liu2019benchmarking}, which
		is an intent classification data set
		containing user utterances of
		$64$ intents;
		\item \textit{Liu54}~\citep{liu2019benchmarking}, which
		is an imbalanced intent classification data set
		of 54 classes collected
		on Amazon Mechanical Turk.
	\end{enumerate*}
	We use the meta-training/meta-validation/meta-testing splits provided in
	\citet{chen2022contrastnet}.
	A summary of the data sets is in Table \ref{table:ds}.
	
	Following~\citep{bao2020few, han2021meta, chen2022contrastnet, hou2022metaprompting},
	we perform	experiments in the 5-way 1-shot and  5-way 5-shot settings
	with
	$15$ query samples per class.
	The pre-trained BERT (\textit{bert-base-uncased}) from
	HuggingFace~\citep{wolf2019huggingface}
	is used as
	the pre-trained MLM
	as in~\citep{chen2022contrastnet, hou2022metaprompting}.
	Experiments are run on a DGX station with $8$ V100 $32$GB GPUs.
	The experiment is repeated three times
	with different random seeds.
	
	\begin{table}[!h]
		\vskip -.2in
		\centering
		\caption{Statistics of the data sets.}
		\resizebox{.48\textwidth}{!}{
			\begin{tabular}{c c c c}
				\toprule
				& \#classes & \#samples & \#tokens per sample \\
				& {(meta-train/valid/test)} & &  (mean $\pm$ std) \\
				\midrule
				\textit{20News} & $8/5/7$ & $18, 820$ &  $340 \pm 151$ \\
				\textit{Amazon} & $10/5/9$ & $24, 000$   & $140 \pm 32$ \\
				\textit{HuffPost} &$ 20/5/16 $ & $36, 900$  & $11 \pm 4$ \\
				\textit{Reuters} & $15/5/11$ & $620$  & $168 \pm 136$ \\
				\textit{HWU64} & $23/16/25$ & $11, 036$ & $7 \pm 3$ \\
				\textit{Liu54} & $18/18/18$ & $25, 478$ & $8 \pm 4$\\
				\bottomrule
			\end{tabular}
		}
		\label{table:ds}
		\vskip -.1in
	\end{table}

	\subsection{Evaluation on RepVerb}
	\label{subsec:expt-verb}
	
	\begin{table*}[!ht]
		\centering
		\vskip -.1in
		\caption{Meta-testing accuracy of various verbalizers on 5-way few-shot classification.
		} \label{table:verb}
		\resizebox{.98\textwidth}{!}{
			\begin{tabular}{ c c c c c c c c}
				\toprule
				& &\textit{20News} & \textit{Amazon}& \textit{HuffPost}& \textit{Reuters} & \textit{HWU64} & \textit{Liu54}  \\
				\midrule
				\multirow{3}{*}{\adjustbox{stack=ll}{$5$-shot}}
				&WARP~\citep{hambardzumyan2021warp} &  $61.43 \pm 0. 15$ & 	$59.53 \pm 0.20$ & $46.31 \pm 0.31$ & $68.67 \pm 0.71$ & $68.60 \pm 0.40$ & $73.11 \pm 0.26$ \\
				& ProtoVerb~\citep{cui2022prototypical} &    $71.33 \pm 0.11$ & 	$71.74 \pm 0.21$ & $57.93 \pm 0.17$ & $80.93 \pm 0.54$ & $73.43 \pm 0.51$ & $76.19 \pm 0.33$  \\
				& RepVerb  &$\mathbf{78.81} \pm 0.08$ & $\mathbf{77.56} \pm 0.16$ &  $\mathbf{61.90} \pm 0.08$ & $\mathbf{88.33} \pm 0.40$ & $\mathbf{78.37 }\pm 0.49$ & $\mathbf{82.14} \pm 0.23$\\
				\midrule
				\multirow{3}{*}{$1$-shot } & WARP~\citep{hambardzumyan2021warp}  &   $49.87 \pm 0. 63$ & 	$48.94 \pm 0.34$ & $38.21 \pm 0.35$ & $52.88 \pm 0.67$ & $53.20 \pm 0.76$ & $58.68 \pm 0.64$ \\
				& ProtoVerb~\citep{cui2022prototypical} &   $54.13 \pm 0.46$ & 	$55.07 \pm 0.27$ & $41.40 \pm 0.21$ & $57.27 \pm 0.73$ & $55.17 \pm 0.81$ & $60.16 \pm 0.37$ \\
				& RepVerb  &  $\mathbf{59.86} \pm 0.38$ & $\mathbf{59.18} \pm 0.31$ &  $\mathbf{44.65} \pm 0.20$ & $\mathbf{63.63} \pm 0.41$ & $\mathbf{59.83} \pm 0.71$ & $\mathbf{66.17} \pm 0.40$\\
				\bottomrule
			\end{tabular}
		}
		\vskip -.05in	
	\end{table*}
	
	First, we compare the performance of the proposed RepVerb with state-of-the-art
	soft verbalizers:
	\begin{enumerate*}[(i), series = tobecont, itemjoin = \quad]
		\item WARP~\citep{hambardzumyan2021warp}\footnote{Note that the verbalizer
		of WARP is the same as that of DART~\citep{zhang2022differentiable}. Its
		implementation is described in Appendix \ref{sec:app-warp}.},
		and 
		\item ProtoVerb~\citep{cui2022prototypical}.
		As the focus is on evaluating verbalizers,
		all methods use the same discrete prompt
		``\tttext{Topic is [MASK]}'', and fine-tune
		all parameters for $5$ steps with a learning rate of $0.00005$ as in \citet{cui2022prototypical}.
	\end{enumerate*}
	
	\textbf{Results}.
	Table \ref{table:verb}
	reports the meta-testing accuracies.
	As can be seen,
	RepVerb outperforms
	WARP and ProtoVerb
	on both  the $1$-shot and  $5$-shot settings.

\begin{figure}[!h]
		\centering
		\vskip -.2in
		\!\!\!
		\subfigure[WARP.\label{fig:warp-Reuters}]{\includegraphics[width=0.166\textwidth]{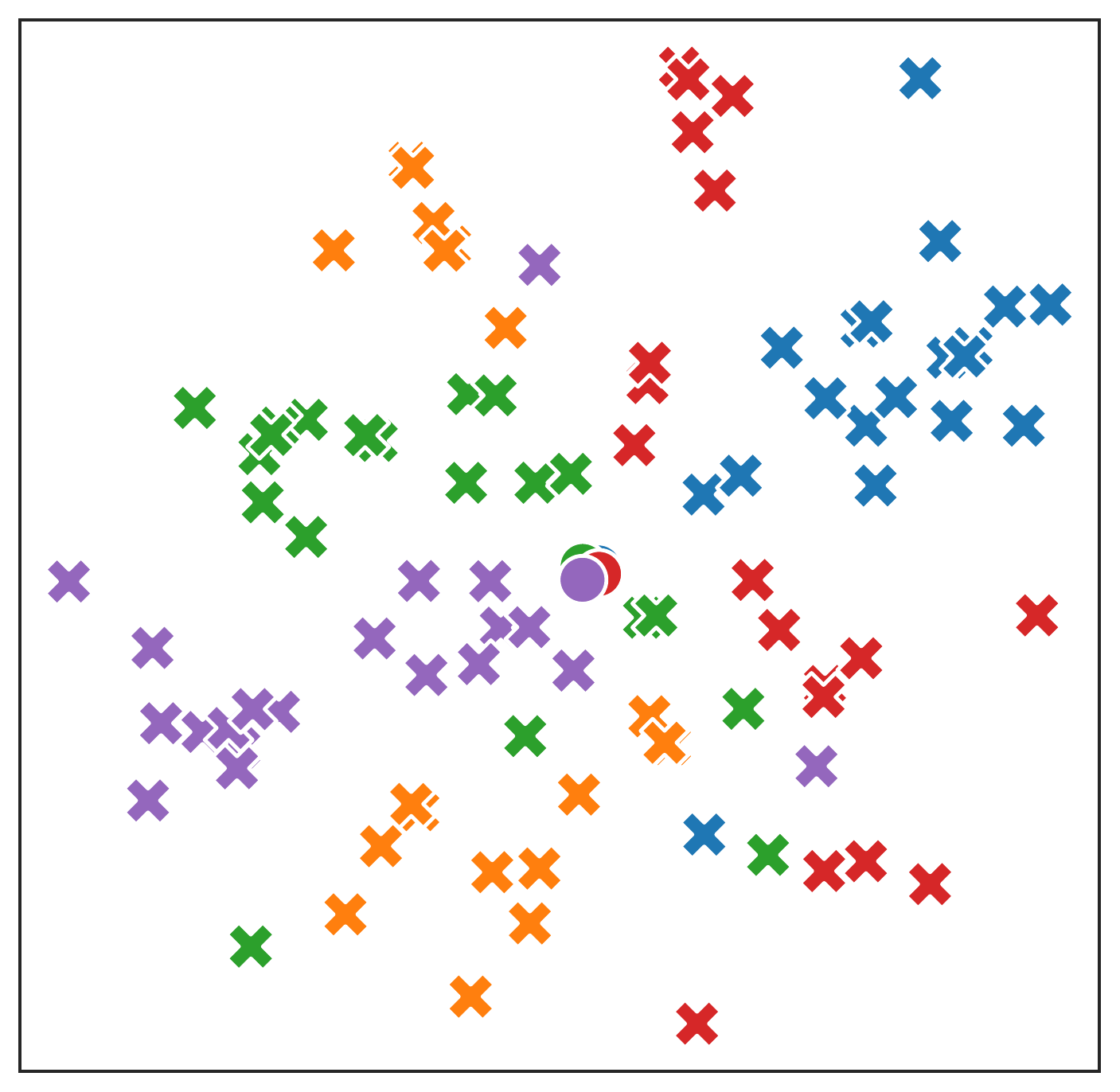}} \!\!\!
		\subfigure[ProtoVerb.\label{fig:protoverb-Reuters}]{\includegraphics[width=0.166\textwidth]{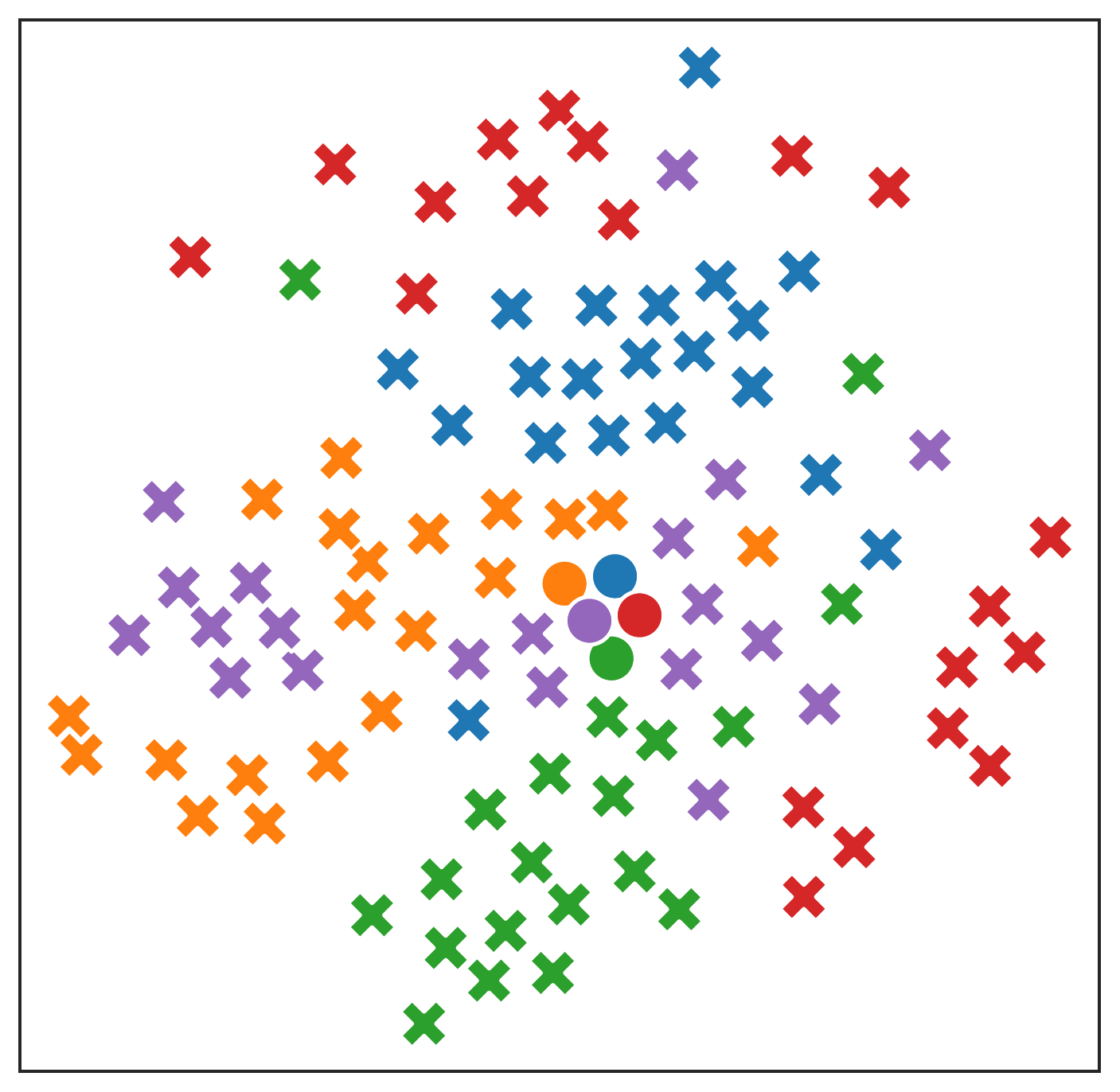}}\!\!
		\subfigure[RepVerb.\label{fig:repverb-Reuters}]{\includegraphics[width=0.166\textwidth]{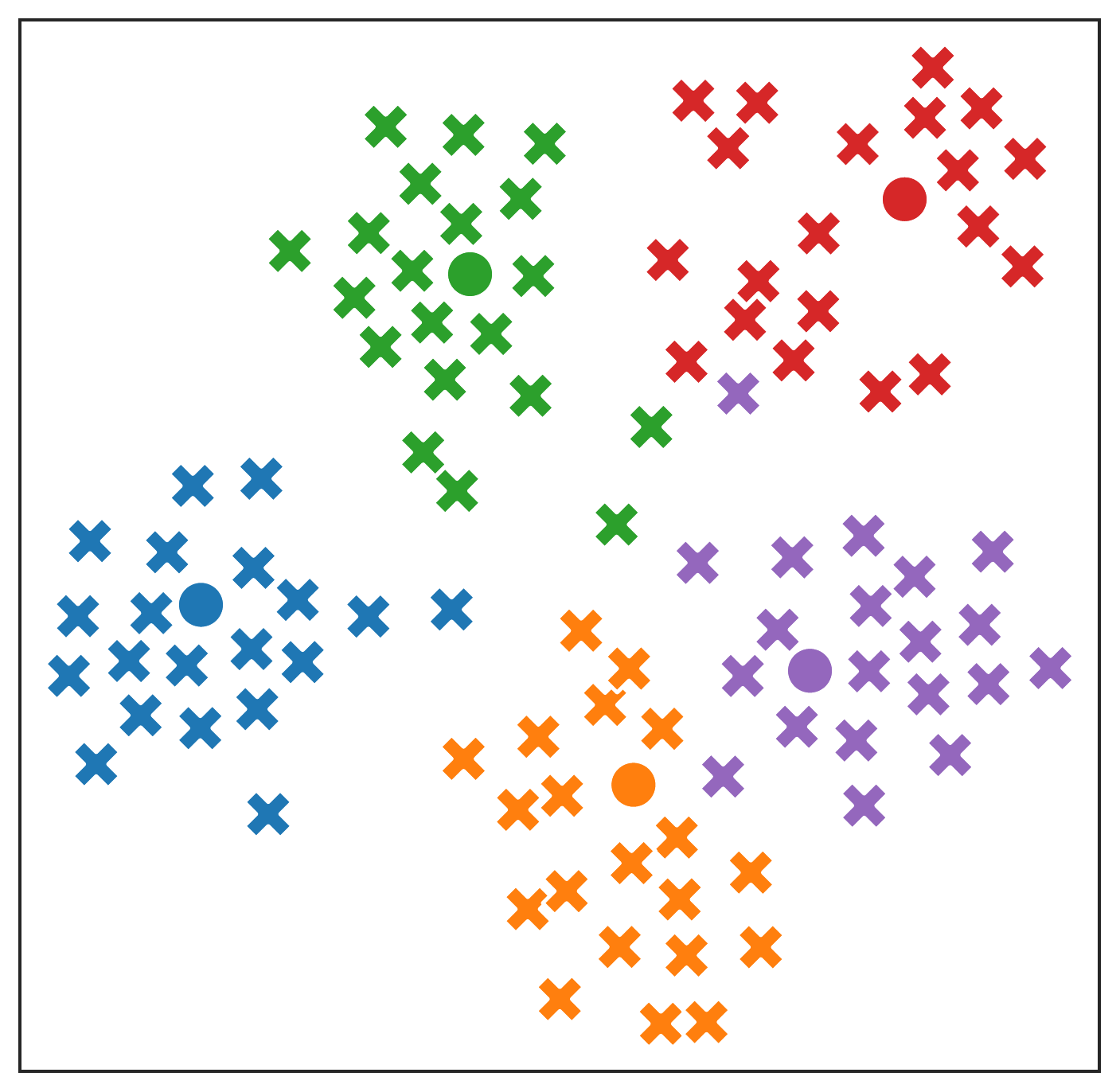}}\!\!\!
		\vskip -.2in
		\caption{t-SNE visualization of \tttext{[MASK]}'s embeddings (crosses) and
			label embeddings (circles) for a 5-way 5-shot task randomly sampled from \textit{Reuters}.}
		\label{fig:visualize-Reuters}
		\vskip -.2in
	\end{figure}

	Figure \ref{fig:visualize-Reuters} shows the t-SNE visualization of the embeddings ($\vh_{\tttext{[MASK]}}(\vx)$'s) of $100$ samples ($\vx$'s)\footnote{5-way $\times$ (5 support samples + 15 query samples) = 100.} and learned label embeddings ($\vv_y$'s)
	for a random 5-way 5-shot task from \textit{Reuters}.\footnote{Results on the other data sets are in Figure~\ref{fig:visualize-others} of Appendix \ref{sec:app-verb}.}
	As can be seen, the
	RepVerb embedding is more discriminative and compact
	than those of WARP and ProtoVerb.
	Moreover, by design, RepVerb's label embedding is consistent with the samples'
	feature embeddings, while those of
	WARP and ProtoVerb are not.
	
	\begin{table*}[!h]
		\vskip -.1in
		\centering
		\caption{Number of parameters and 
			5-way 5-shot  classification
			meta-testing accuracy.
			Results marked with $^\dagger$ are from \citet{chen2022contrastnet}.
			``--''
			indicates that the corresponding result
			is not reported in \citet{chen2022contrastnet}.
		} \label{table:main-5-shot}
		\resizebox{.98\textwidth}{!}{
			\begin{tabular}{c c c c c c c c }
				\toprule
				&\#param $(\times 10^6)$ &  \textit{20News} &\textit{Amazon}& \textit{HuffPost}&  \textit{Reuters} & \textit{HWU64} & \textit{Liu54} \\
				\midrule
				HATT$^\dagger$~\citep{gao2019hybrid} & $0.07$ & $55.00$ & $66.00$  & $56.30$ & $56.20$ &-&- \\
				DS$^\dagger$~\citep{bao2020few} &$1.73$& $68.30$ & $81.10$ & $63.50$ & $96.00$ &-&-\\
				MLADA$^\dagger$~\citep{han2021meta} &$0.73$ & $77.80$ & $86.00$ & $64.90$ & $96.70$ &-&-\\
				ContrastNet$^\dagger$~\citep{chen2022contrastnet} &$109.52$& $71.74$ & $85.17$ & $65.32$ & $95.33$ &$92.57$& $93.72$ \\
				\midrule
				MetaPrompting~\citep{hou2022metaprompting} &$109.52$ &  $85.67 \pm 0.44 $  & $84.19 \pm 0.30$ & $72.85\pm 1.01$ & $95.89\pm 0.23$& $93.86 \pm 0.97$& $94.01 \pm 0.26$
				\\
				MetaPrompting+WARP &$109.52$ & 	$85.81\pm 0.48$&$85.54 \pm 0.20$&$71.71 \pm 0.72$&	$97.28 \pm 0.30$&$93.99 \pm 0.76$&$94.33 \pm 0.27$\\
				MetaPrompting+ProtoVerb &$109.52$ &  $86.18\pm 0.51$ & $84.91 \pm 0.38$ & $73.11 \pm 0.80$ & $97.24 \pm 0.25$ & $93.81 \pm 0.81$ & $94.38 \pm 0.18$\\
				MetaPrompting+RepVerb &$109.52$&  $86.89\pm 0.39$ & $85.98 \pm 0.28$& $74.62\pm 0.88$ & $97.32  \pm 0.31$& $94.23 \pm 0.67$
				&
				$94.45 \pm 0.33$ \\
				MetaPrompter   &$0.06$& $\mathbf{88.57}\pm 0.38$& $\mathbf{86.36} \pm 0.24$ & $\mathbf{74.89} \pm 0.75$ & $\mathbf{97.63} \pm 0.22$&$\mathbf{ 95.30}\pm 0.51$& $\mathbf{95.47} \pm 0.21$
				\\
				\bottomrule
		\end{tabular}		}
		\vskip -.05in
	\end{table*}
	
	\begin{table*}[!h]
		\centering
		\vskip -.15in
		\caption{Number of parameters and 5-way 1-shot  Meta-testing classification accuracy.
			Results marked with $^\dagger$ are from \citet{chen2022contrastnet}.
			``--''
			indicates that the corresponding result
			is not reported in \citet{chen2022contrastnet}.
		} \label{table:main-1-shot}
		\resizebox{.98\textwidth}{!}{
			\begin{tabular}{ c c c c c c c c }
				\toprule
				& \#param $(\times 10^6)$ &\textit{20News} & \textit{Amazon}& \textit{HuffPost}& \textit{Reuters} & \textit{HWU64} & \textit{Liu54} \\
				\midrule
				HATT$^\dagger$~\citep{gao2019hybrid} & $0.07$& $44.20$ & $49.10$ & $41.10$ & $43.20$  &-& -\\
				DS$^\dagger$~\citep{bao2020few} &$1.73$ &  $52.10$ & $62.60$ & $43.00$ & $81.80$&-& - \\
				MLADA$^\dagger$~\citep{han2021meta} & $0.73$& $59.60$ & $68.40$ & $64.90$ & $82.30$ &-& - \\
				ContrastNet$^\dagger$~\citep{chen2022contrastnet} & $109.52$  & $71.74$ & $76.13$ & $53.06$ & $86.42$ &$86.56$&   $85.89$ \\
				\midrule
				MetaPrompting~\citep{hou2022metaprompting}  &$109.52$&   $82.46\pm 0.50$  & $76.92 \pm 0.77$  & $68.62 \pm 0.56$ & 	$92.56 \pm 0.77$ &$91.06 \pm 0.41$&$87.79 \pm 0.29$ \\
				MetaPrompting +WARP&$109.52$& $82.93 \pm 0.39$& $78.27 \pm 0.72$ & $67.78 \pm 0.41$ & $94.74 \pm 0.56$ & $91.30 \pm 0.35$ & $88.69 \pm 0.26$  \\
				MetaPrompting+ProtoVerb &$109.52$&  $83.15 \pm 0.41$ & $78.19\pm 0.65$ & $68.96 \pm 0.52$ & $95.26\pm 0.40$ & $91.27 \pm 0.63$ & $90.05 \pm 0.15$\\
				MetaPrompting+RepVerb &$109.52$& $84.13 \pm 0.30$ & $78.59\pm 0.43$ & $\mathbf{69.02}\pm 0.51$ & $95.78\pm 0.33$ & $91.32\pm 0.44$ & $90.13 \pm 0.20$ \\
				MetaPrompter  & $0.06$ &   $\mathbf{84.62} \pm 0.29$ &$\mathbf{79.05}\pm 0.21$ & $67.12\pm 0.23$ & $\mathbf{96.34} \pm 0.20$&$\mathbf{92.11} \pm 0.30$&$\mathbf{93.72} \pm 0.18$
				\\
				\bottomrule
			\end{tabular}
		}
		\vskip -.05in
	\end{table*}

	\subsection{Evaluation on MetaPrompter}
	\label{sec:eval-MetaPrompter}

	We compare MetaPrompter with a variety of
	baselines. These include
	state-of-the-art prompt-based methods of
	\begin{enumerate*}[(i), series = tobecont, itemjoin = \quad]
		\item MetaPrompting~\citep{hou2022metaprompting},
		and its variants
		\item
		MetaPrompting+WARP~/~MetaPrompting+ProtoVerb~/~MetaPrompting+RepVerb,
		which combine MetaPrompting 
		with the soft verbalizer of WARP / ProtoVerb / RepVerb, respectively.
		Moreover, we also compare with the
		non-prompt-based methods of:
		\item HATT~\citep{gao2019hybrid}, which
		meta-learns a
		prototypical network~\citep{snell2017prototypical} with a hybrid attention mechanism;
		\item DS~\citep{bao2020few}, which learns attention scores based on word frequency;
		\item MLADA~\citep{han2021meta}, which uses an
		adversarial domain adaptation network
		to extract domain-invariant features
		during meta-training; and
		\item ContrastNet~\citep{chen2022contrastnet}, which performs feature
		extraction by
		contrastive learning.
	\end{enumerate*}
	
	For MetaPrompter, hyperparameters $K$ and $L_p$
	are chosen from $\{1,2,4,8, 16, 32, 64\}$
	using the meta-validation set.
	For the base learner,
	$\alpha = 0.1$, and
	$J=5$ (resp. $15$) at meta-training (resp. meta-validation or meta-testing).
	We train the prompt pool
	for $T = 3, 000$ iterations using the Adam optimizer~\citep{kingma2015adam} with a learning rate of $0.001$.
	To prevent overfitting, we evaluate the meta-validation
	performance every $50$ iteration and choose the checkpoint with the best meta-validation performance for meta-testing.
	For the hand-crafted verbalizer
	used in \eqref{eq:hard-pred},
	label tokens are obtained by tokenizing the class name and its synonyms as in \citep{hou2022metaprompting, hu2022knowledgeable}.
	Following~\citet{lester2021power},
	prompts are initialized from input
	embeddings of randomly sampled label tokens for both MetaPrompting and MetaPrompter.

	\textbf{Results}.
	Table \ref{table:main-5-shot}
	shows the
	number of parameters and
	meta-testing accuracy in the
	$5$-shot setting.
	As can be seen,
	MetaPrompter
	is more accurate than both prompt-based
	and non-prompt-based
	baselines.
	Moreover,
	since MetaPrompter
	only tunes the
	prompt pool
	and keeps the language model frozen,
	it has much fewer meta-parameters
	than
	MetaPrompting
	and
	ContrastNet.
	
    Furthermore, MetaPrompting+RepVerb performs better than MetaPrompting+WARP and
    MetaPrompting+ProtoVerb, demonstrating that the proposed RepVerb is also beneficial to MetaPrompting.

	Table \ref{table:main-1-shot}
	shows the number of parameters and meta-testing accuracy in the $5$-way $1$-shot
	setting.
	As can be seen,
	the state-of-the-art prompt-based
	methods  always achieve higher accuracies than
	the non-prompt-based ones.
	Furthermore,
	MetaPrompter performs the best on 5 of the 6 data sets. Besides,
	RepVerb is 
	again
	useful to
	MetaPrompting on all six data sets.
	
	\begin{figure}[!h]
		\centering
		\includegraphics[width=0.25\textwidth]{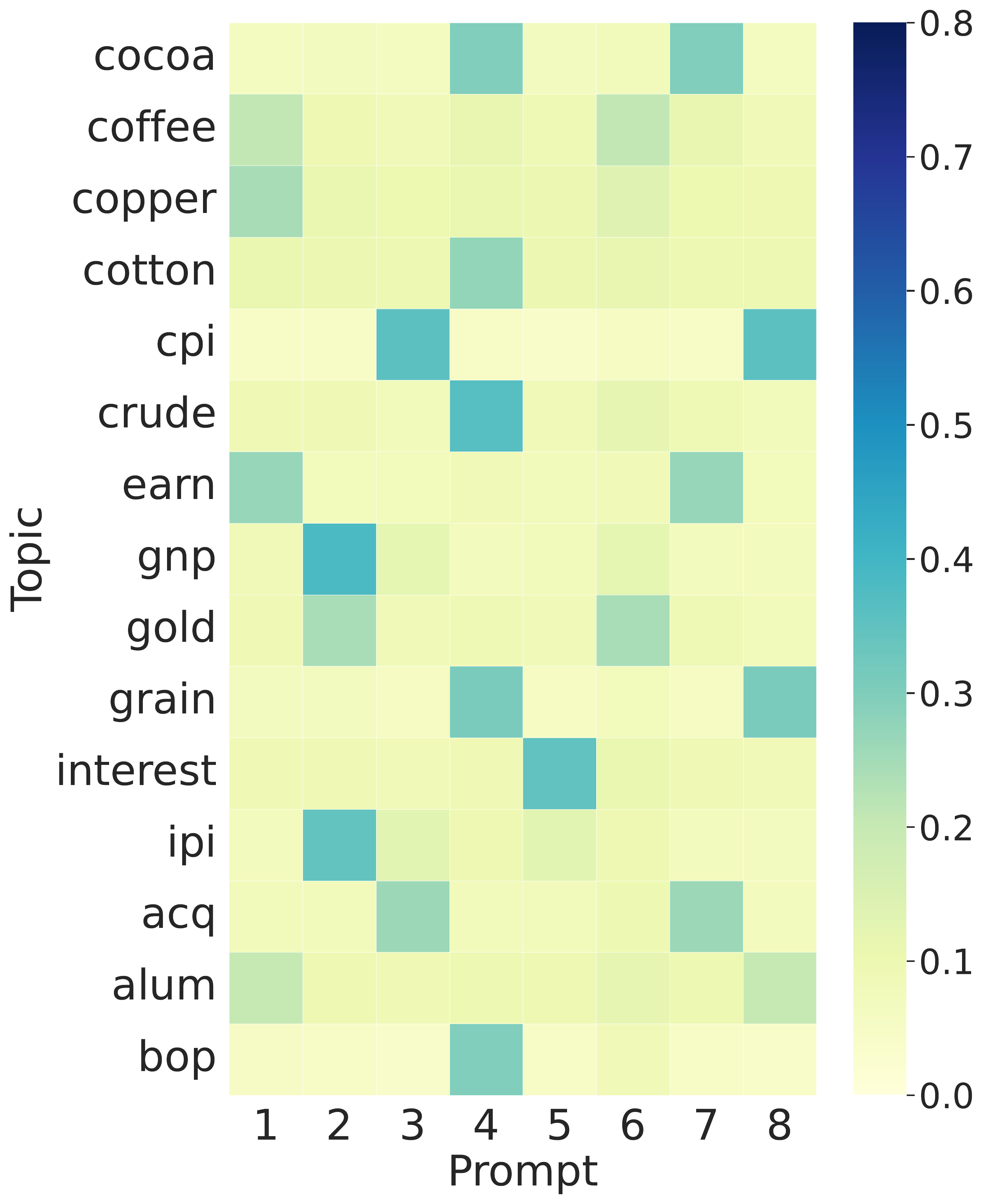}
		\vskip -.15in
		\caption{Distribution of attention weights on 5-way 5-shot classification of \textit{Reuters} ($15$ topics).}
		\label{fig:heatmap-reuster-truncated}
	\end{figure}
	
	\subsection{Visualization}
	
	\begin{table*}[!h]
		\centering
		\caption{Nearest tokens to the learned prompts for \textit{Reuters}.}
		\label{table:nearest-tokens}
		\begin{tabular}{c c}
			\toprule
			prompt id & nearest tokens \\
			\midrule
			1 & copper, steel, trading, gas, fx, aluminum, earn, coffee \\
			2 & gross, ship, index, money, gold, tin, iron, retail\\
			3 & product, cpi, industrial, acquisitions, jobs, supplying, orange, sugar \\
			4 & cocoa, production, grain, livestock, wholesale, cotton, bop, crude \\
			5 & oil, national, rubber, nat, interest, price, reserves, regional \\
			6 & nat, wholesale, sugar, golden, reserves, drinks, production, product \\
			7 & chocolate, sugar, cheat, orange, trade, fx, cash, acquiring\\
			8 & aluminum, livestock, cpc, tin, shops, wheat, petrol, supply \\
			\bottomrule
		\end{tabular}
	\end{table*}
	
	\begin{figure*}[!h]
		\centering
		\includegraphics[width=0.98\textwidth]{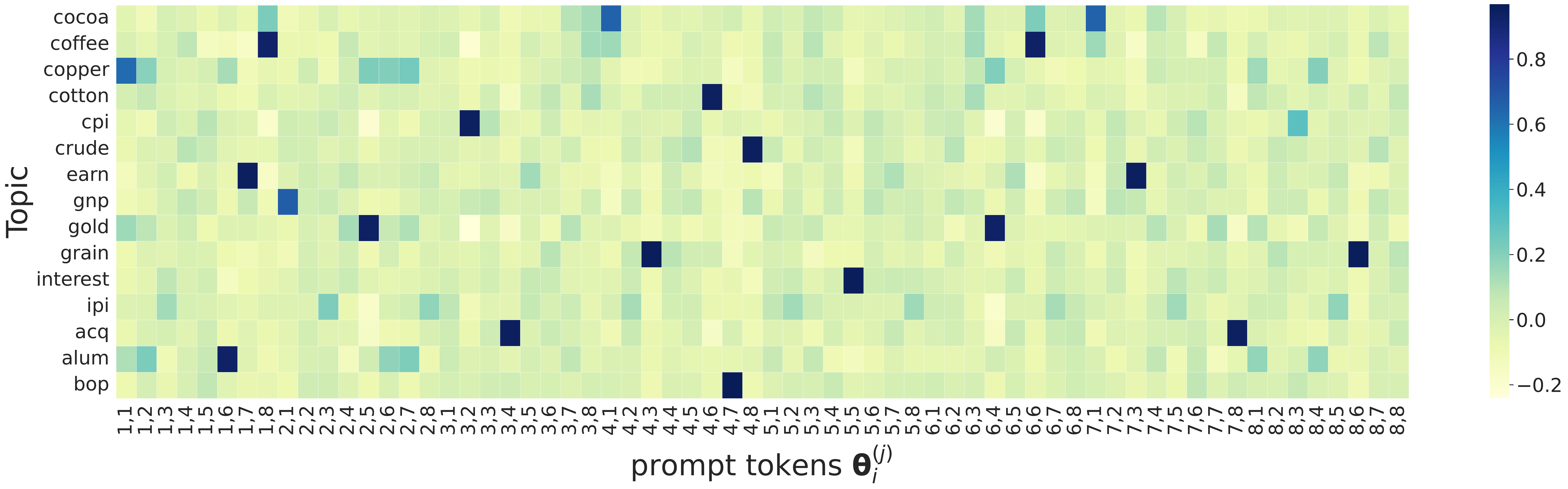}
		\vskip -.2in
		\caption{Cosine similarities between learned prompt tokens and topic embeddings on 5-way 5-shot classification of
			\textit{Reuters}. In the x-axis, $(i,j)$ stands for the $j$th row of $\vtheta_i$ (i.e., $\vtheta_i^{(j)}$)}
		\label{fig:heatmap-reuster-ij-truncated}
		\vskip -.05in
	\end{figure*}

    In this section, we visualize the 
    meta-knowledge	
    in the prompt pool 
    learned 
    from the 5-way 5-shot classification task on 
	\textit{Reuters}.
Table~\ref{table:nearest-tokens} shows the nearest tokens to each of the $K$ 
	($=8$) 
learned prompts.
	Figure \ref{fig:heatmap-reuster-truncated} shows
	the average attention
	weights
	between the $K$ prompts and
	meta-training
	samples
	belonging to class (topic) $y$:
	\[	\frac{1}{|\hT_y|} \sum_{\tau \in \hT_y} \frac{1}{|\hS_{\tau, y}|} \sum_{(\vx, y) \in \hS_{\tau,y}} \softmax\left(\frac{\vK_{T,J}\vq_\vx}{\sqrt{d_o}}\right),\]
	\vskip -.2in
	where $\hT_y$ is the subset of tasks in $\hT$ having class $y$.
As can be seen,
	samples from each target class prefer prompts 
	whose tokens are
	related to that class.
	For example,
	samples from the topic \textit{cocoa}
	tend to use the 4th and 7th prompts 
	(whose tokens are close to words like
	\textit{cocoa}, \textit{chocolate} as can be seen from Table~\ref{table:nearest-tokens}),
	while samples
	from the topic \textit{coffee}
	tend to use the 1st and 6th prompts 
	(whose tokens are close to words like
	\textit{coffee} and \textit{sugar}.

	Recall that the prompt pool has $K$
	learnable prompts
	$\{(\vk_i, \vtheta_i): i=1,\dots, K \}$, with
	key $\vk_i\in \bR^{d_o}$
	and
	value $\vtheta_i \in \bR^{L_p\times d_i}$.
	Let $\vtheta_{i}^{(j)}$ be the $j$th row of $\vtheta_{i}$. 
	Moreover, let 
	$\frac{1}{|\hV_y|} \sum_{\tttext{w}\in \hV_y}\hE(\tttext{w})$ be
	the 
	embedding
	of topic (class) $y$,
	where 
	$\hV_y$
	is a set
	of 
	tokens 
	relevant  to
	label $y$ (obtained from \citet{hou2022metaprompting}),
	and $\hE(\cdot)$ is the input embedding.
	Figure \ref{fig:heatmap-reuster-ij-truncated} shows 
	the cosine similarities between the learned prompt tokens
	$\{\vtheta_{i}^{(j)}:i=1,\dots, K, j=1\dots, L_p\}$ 
	and topic embeddings.
	As can be seen,
	embedding of \textit{cocoa}
	is close to $\vtheta_{4}^{(1)}$ and $\vtheta_{7}^{(1)}$.
	Thus, samples from 
	\textit{cocoa}
	prefer
	the 4th and 7th prompts 
	(Figure \ref{fig:heatmap-reuster-truncated}).
	Similarly,
	embedding of 
	\textit{coffee}
	is close to $\vtheta_{1}^{(8)}$ and $\vtheta_{6}^{(6)}$.
	Thus, samples from
	\textit{coffee}
	prefer
	the 1st and 6th prompts 
	(Figure \ref{fig:heatmap-reuster-truncated}).

		\begin{figure*}[!h]
		\centering
		\!\!
		\subfigure[\textit{20News}.\label{fig:ab-News20-K}]{\includegraphics[width=0.166\textwidth]{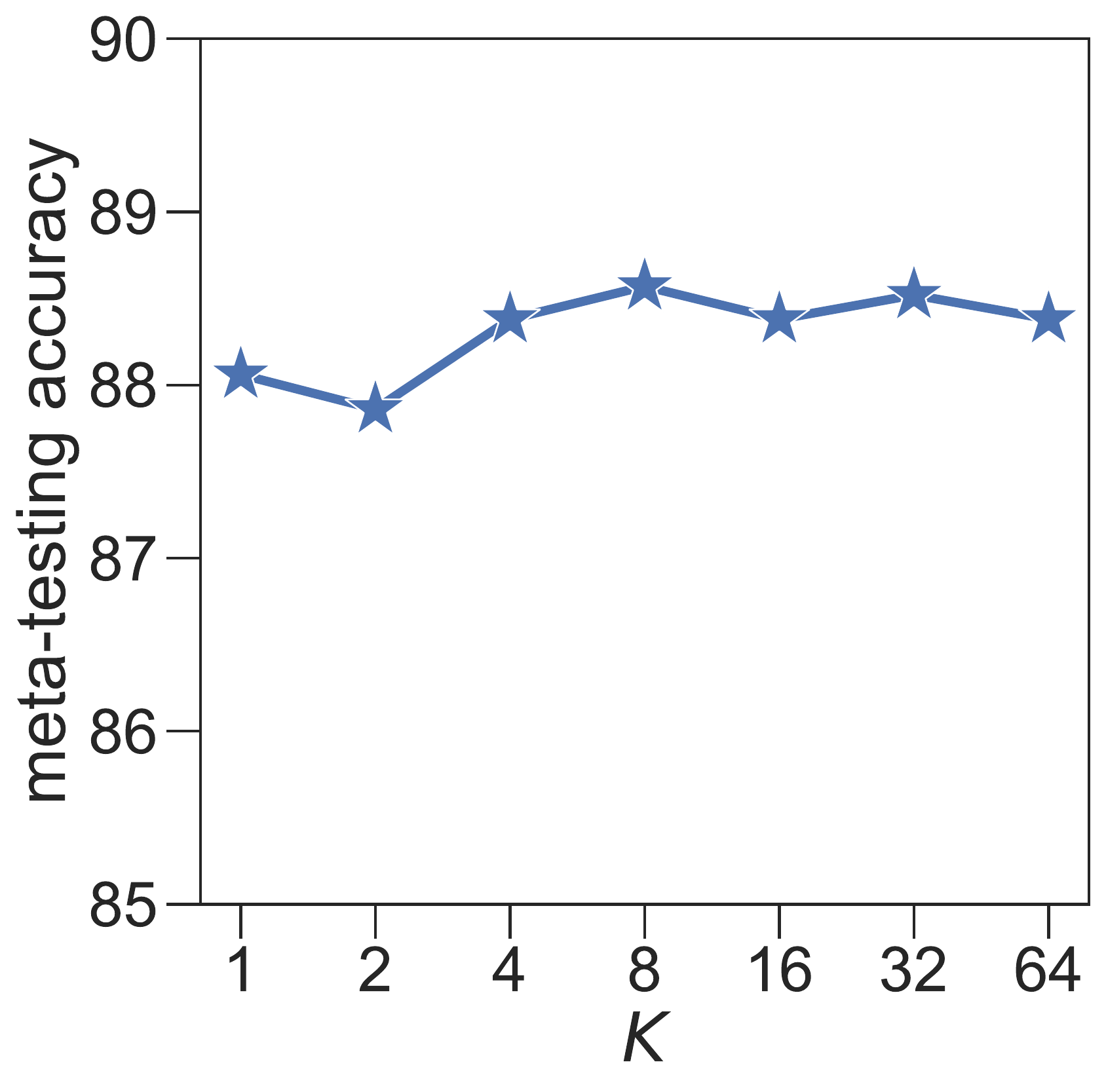}} \!\!
		\subfigure[\textit{Amazon}. \label{fig:ab-Amazon-K}]{\includegraphics[width=0.166\textwidth]{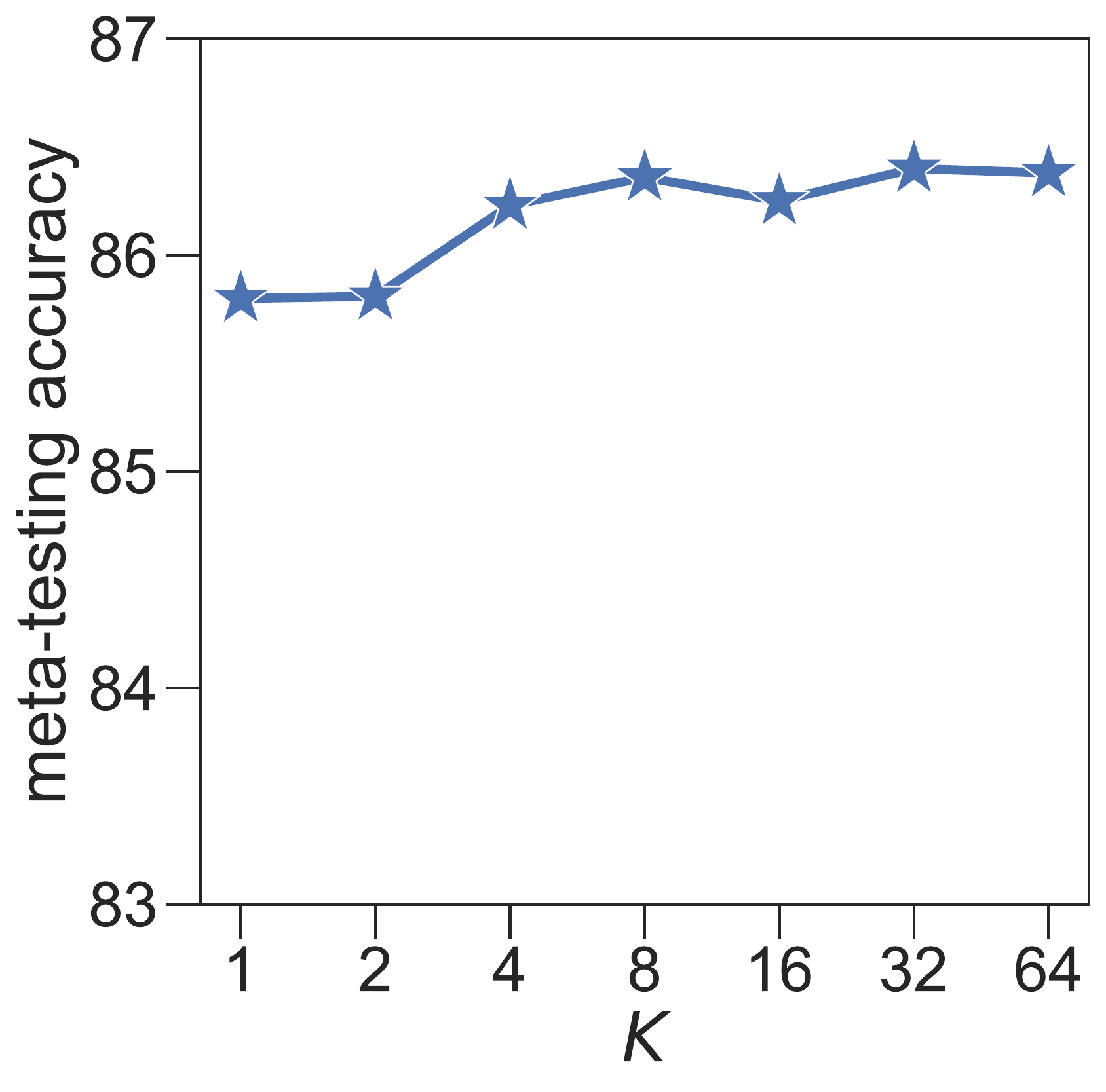}}\!\!
		\subfigure[\textit{HuffPost}.\label{fig:ab-Huffpost-K}]{\includegraphics[width=0.166\textwidth]{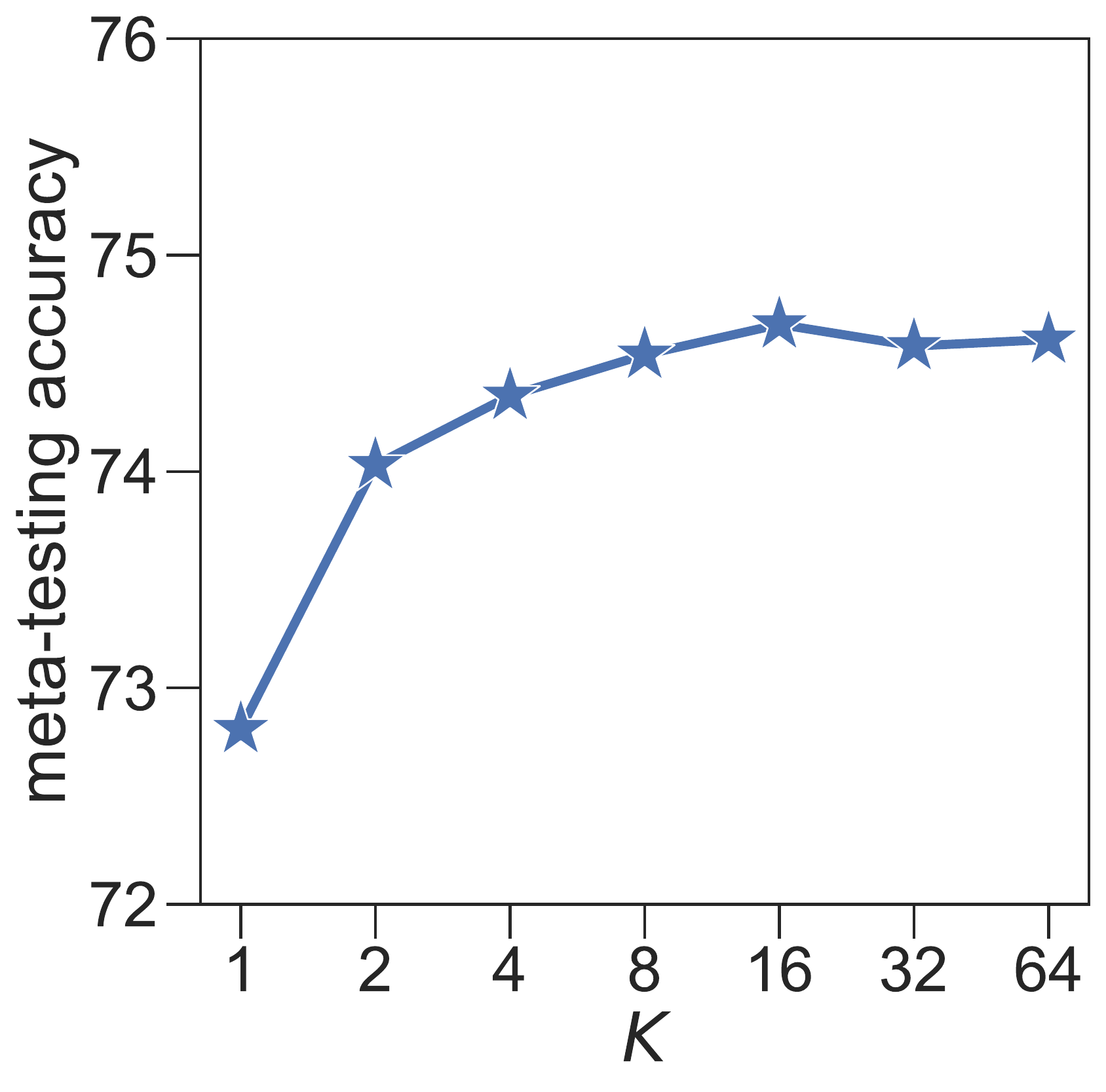}}\!\!
		\subfigure[\textit{Reuters}.\label{fig:ab-Reuters-K}]{\includegraphics[width=0.166\textwidth]{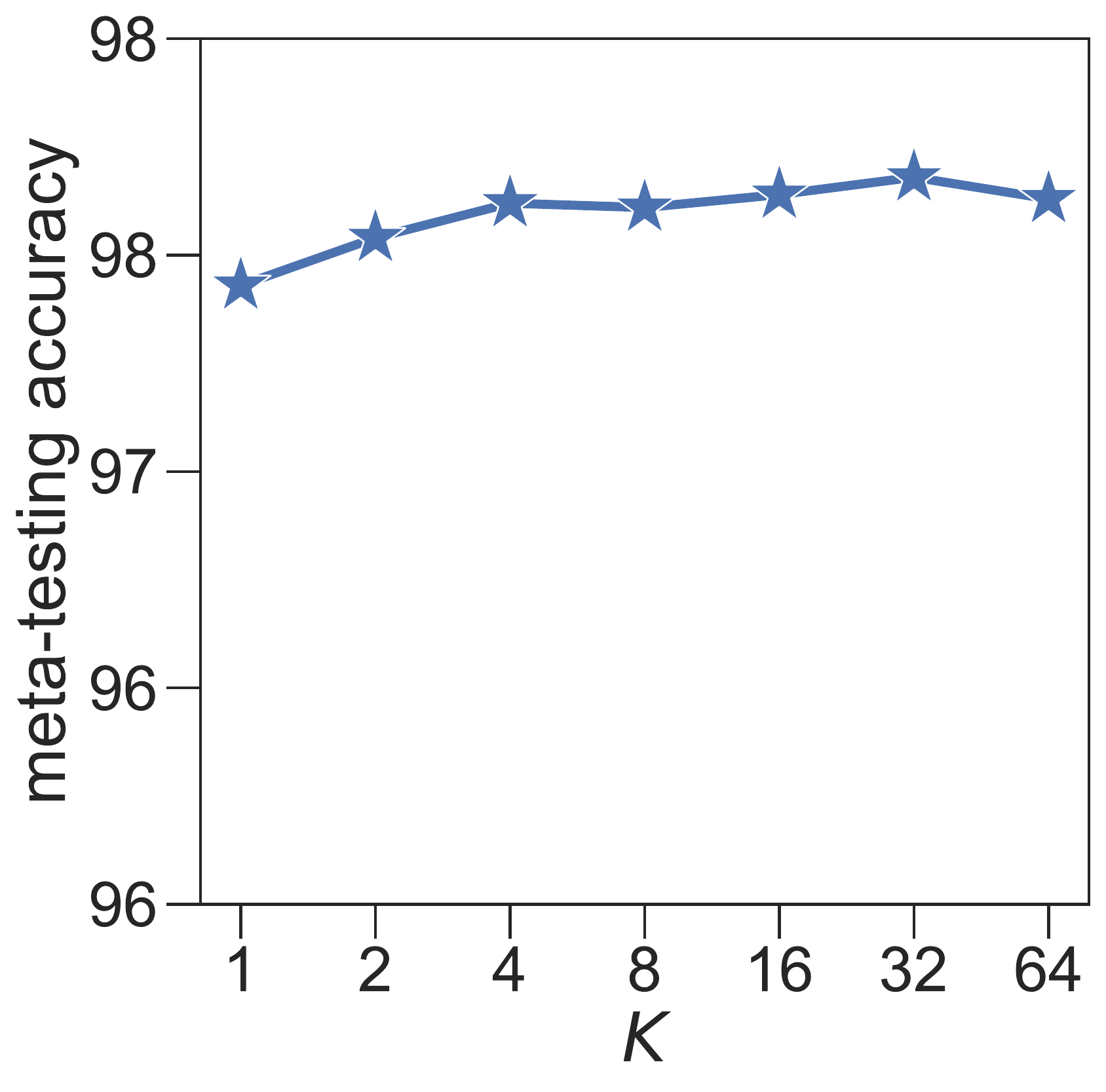}}\!\!
		\subfigure[\textit{HWU64}.\label{fig:ab-hwu-K}]{\includegraphics[width=0.166\textwidth]{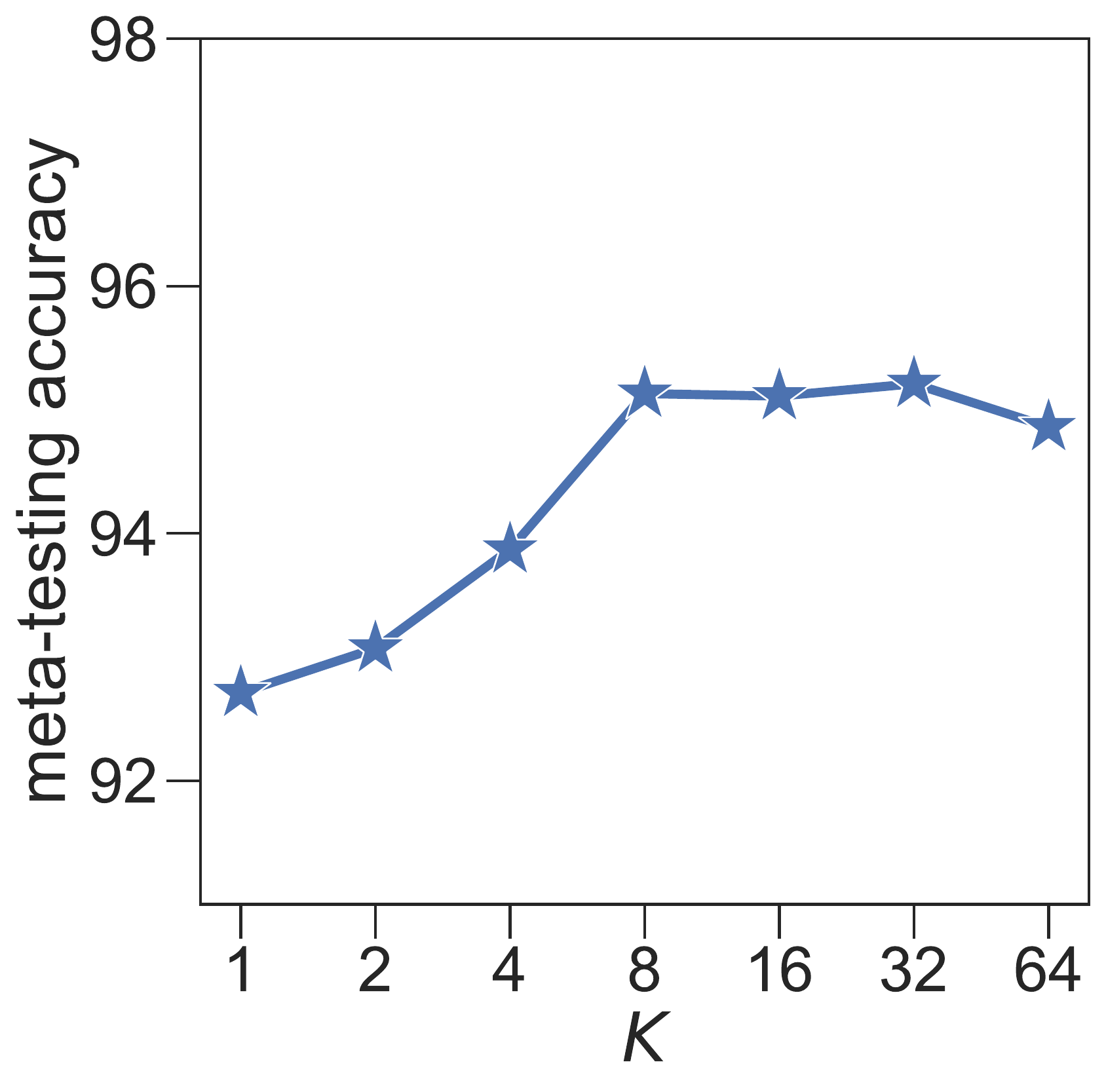}}\!\!
		\subfigure[\textit{Liu54}.\label{fig:ab-liu-K}]{\includegraphics[width=0.166\textwidth]{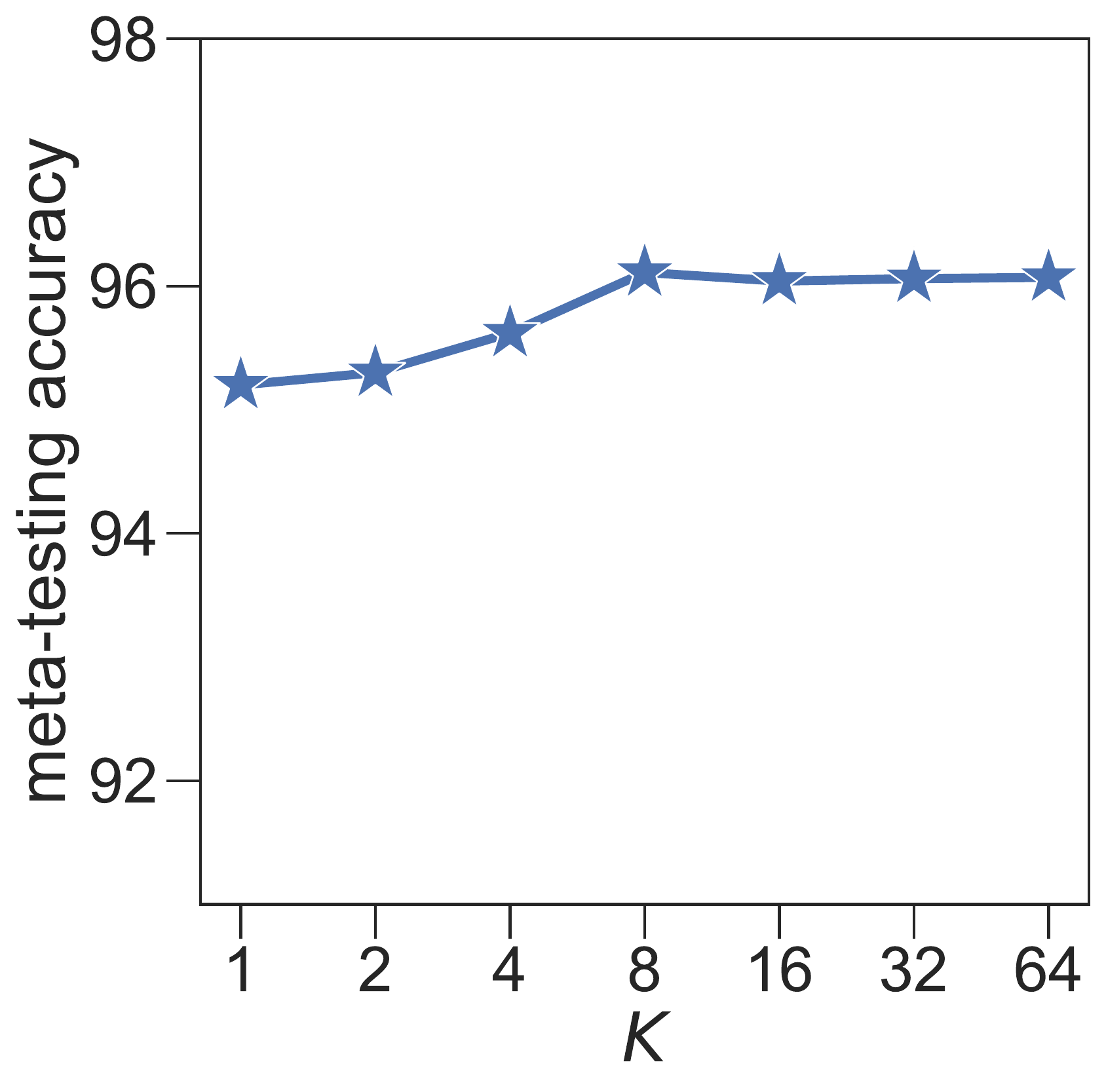}}\!\!
		\vskip -.15in
		\caption{Effect of $K$ (in log-scale) on 5-way 5-shot classification
			($L_p=8$).}
		\label{fig:ab-K}
	\end{figure*}

		\begin{figure*}[!h]
		\centering
		\vskip -.1in
		\!\!
		\subfigure[\textit{20News}.\label{fig:ab-News20-Lp}]{\includegraphics[width=0.166\textwidth]{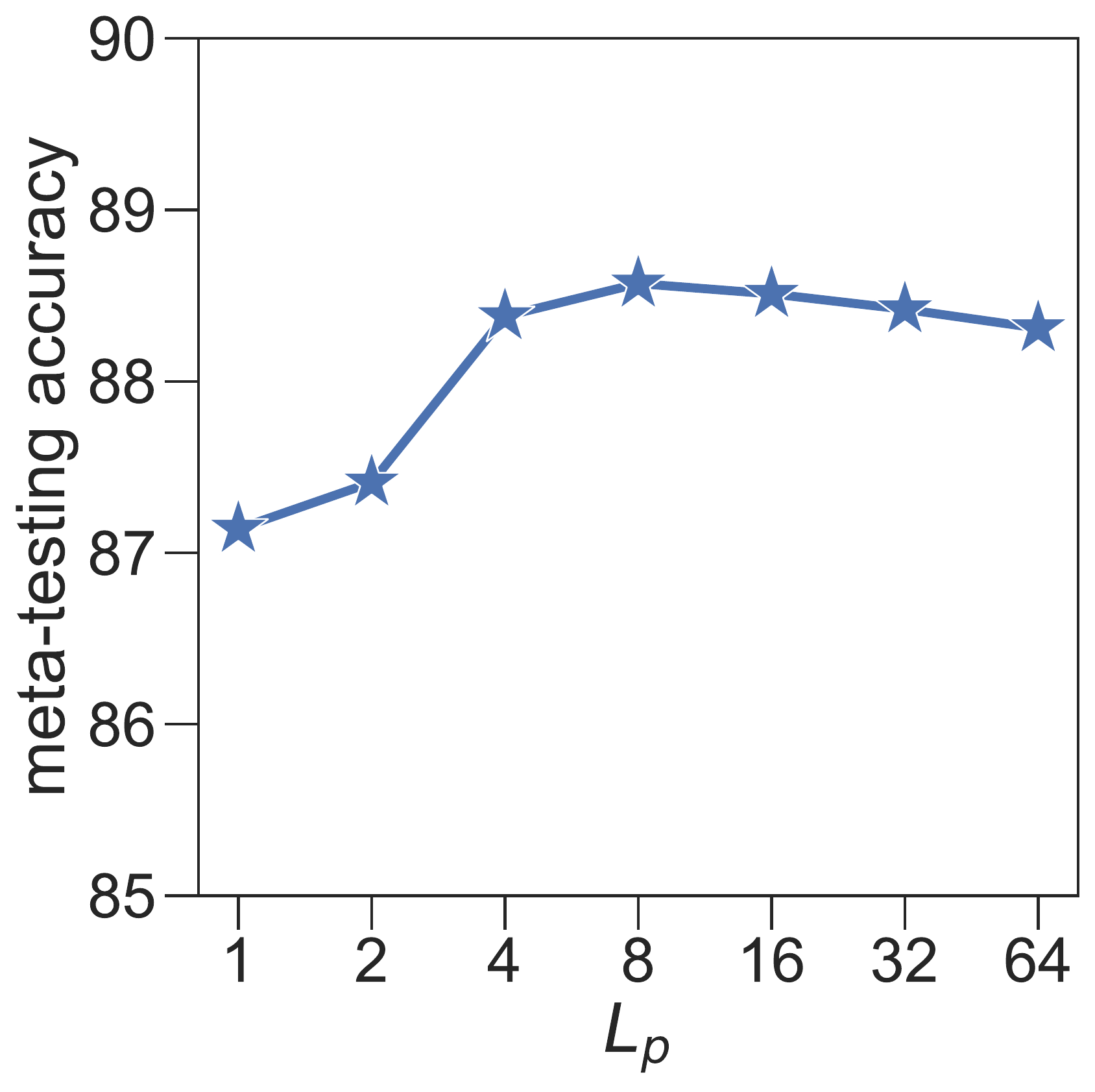}}\!\!
		\subfigure[\textit{Amazon}. \label{fig:ab-Amazon-Lp}]{\includegraphics[width=0.166\textwidth]{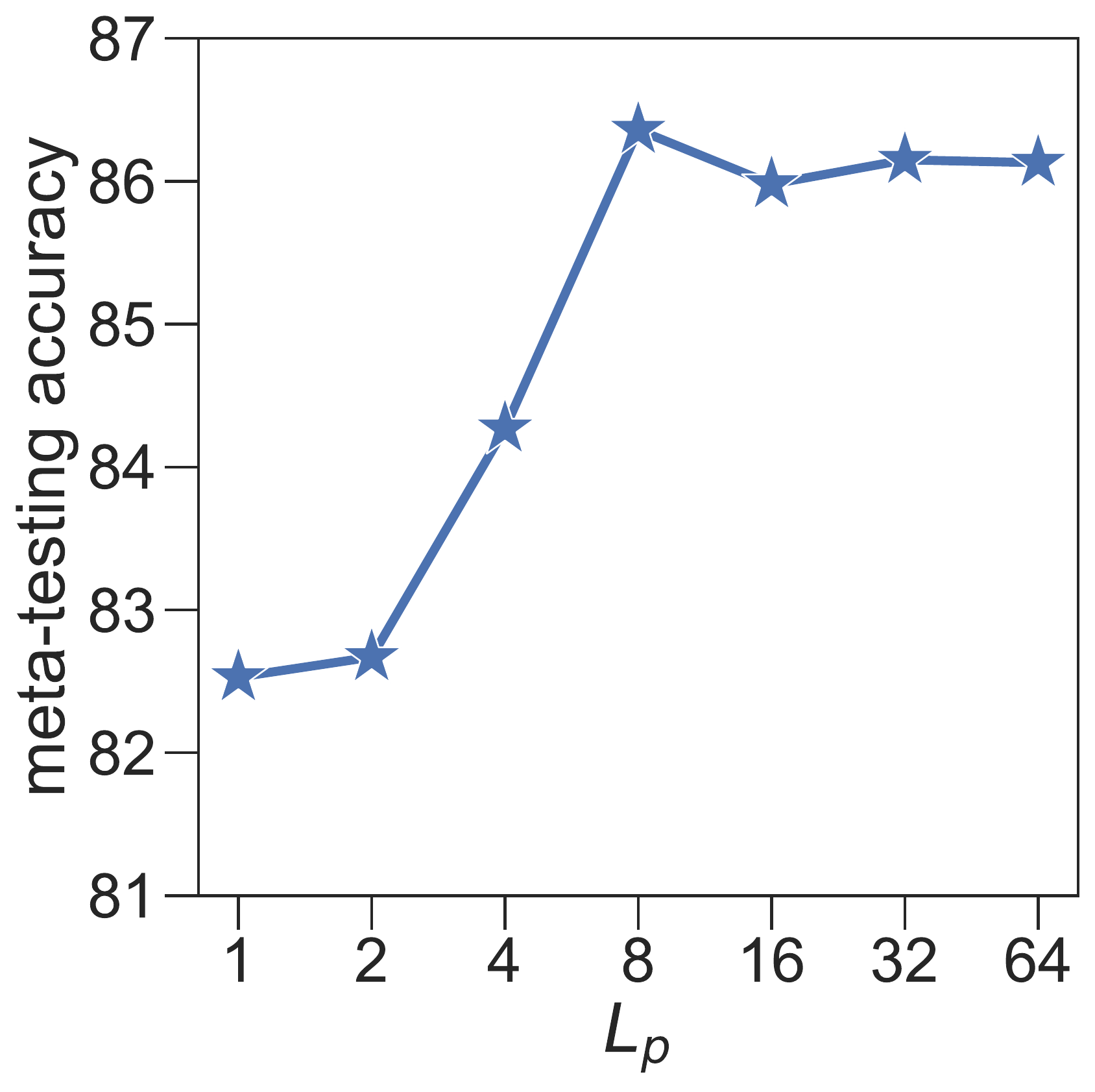}}\!\!
		\subfigure[\textit{HuffPost}.\label{fig:ab-Huffpost-Lp}]{\includegraphics[width=0.166\textwidth]{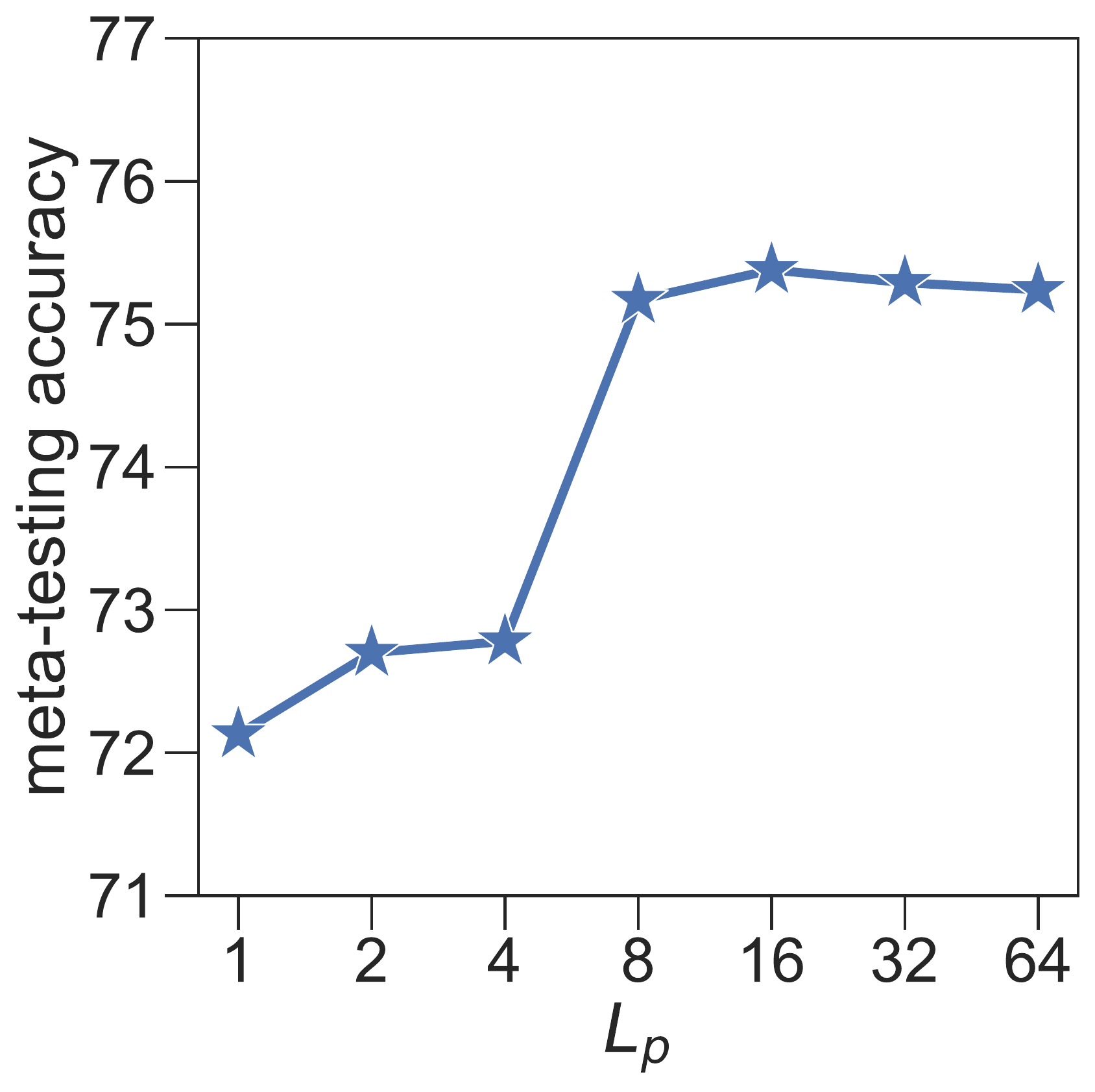}}\!\!
		\subfigure[\textit{Reuters}.\label{fig:ab-Reuters-Lp}]{\includegraphics[width=0.166\textwidth]{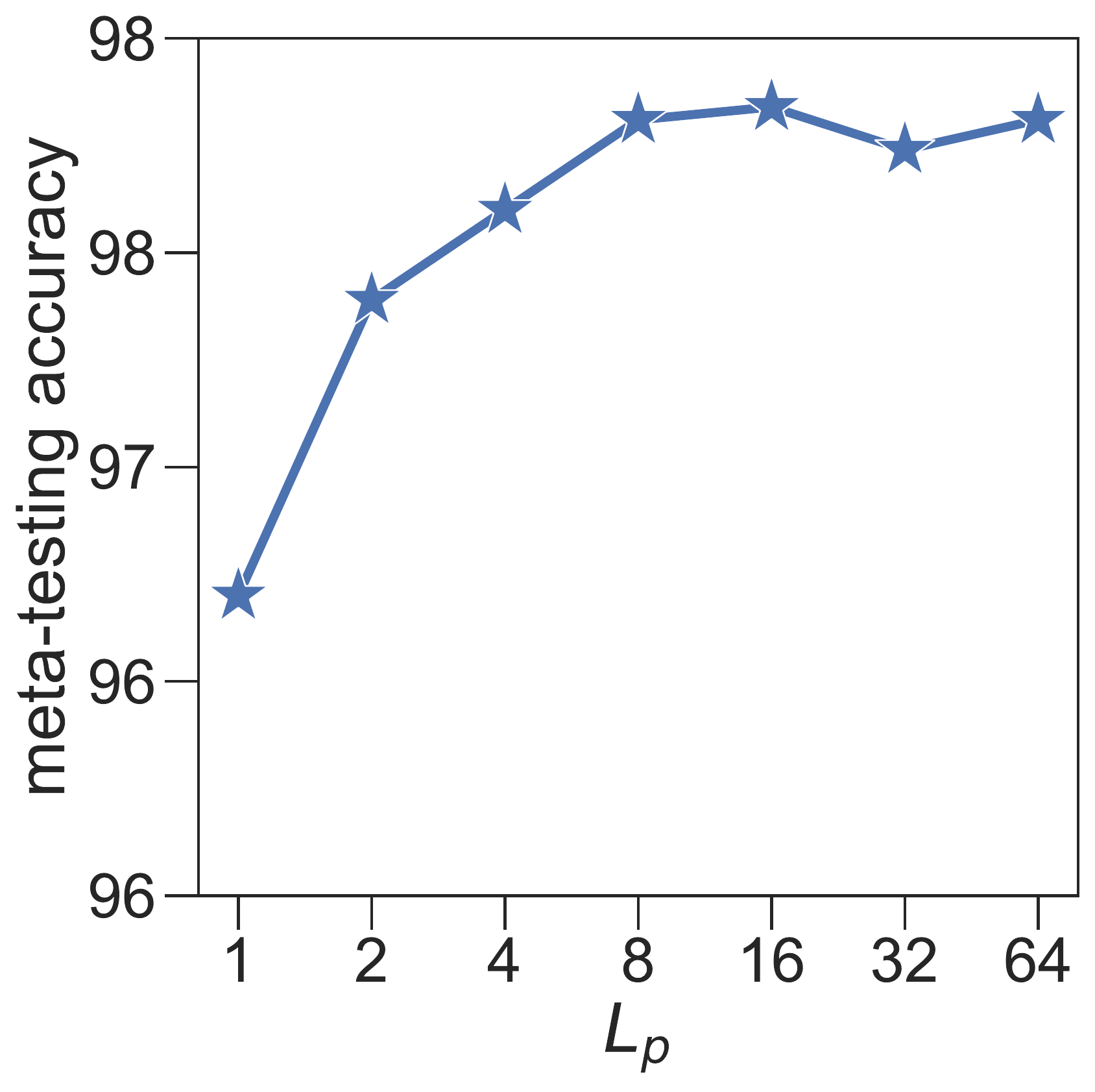}} \!\!
		\subfigure[\textit{HWU64}.\label{fig:ab-hwu-Lp}]{\includegraphics[width=0.166\textwidth]{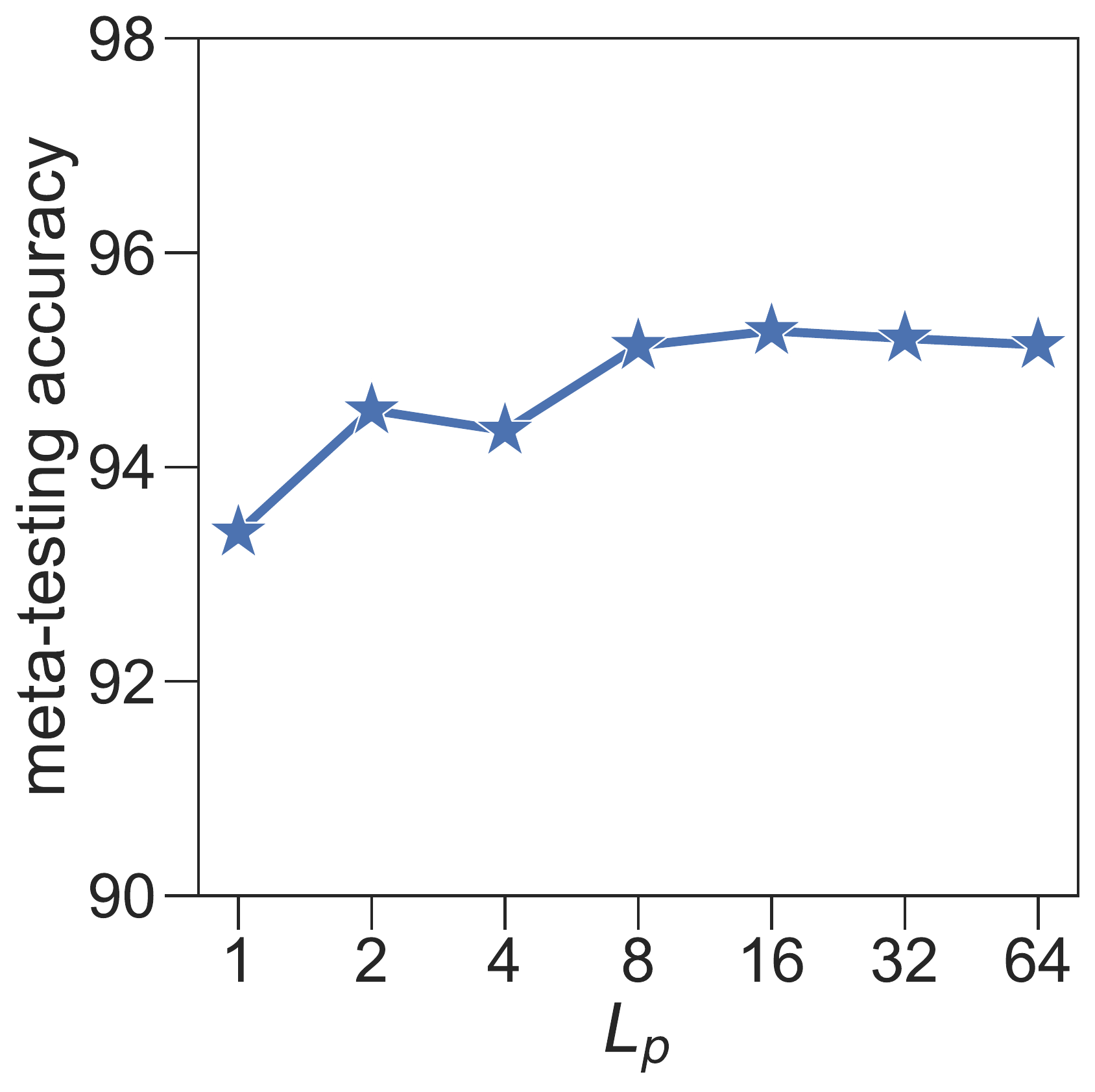}}\!\!
		\subfigure[\textit{Liu54}.\label{fig:ab-liu-Lp}]{\includegraphics[width=0.166\textwidth]{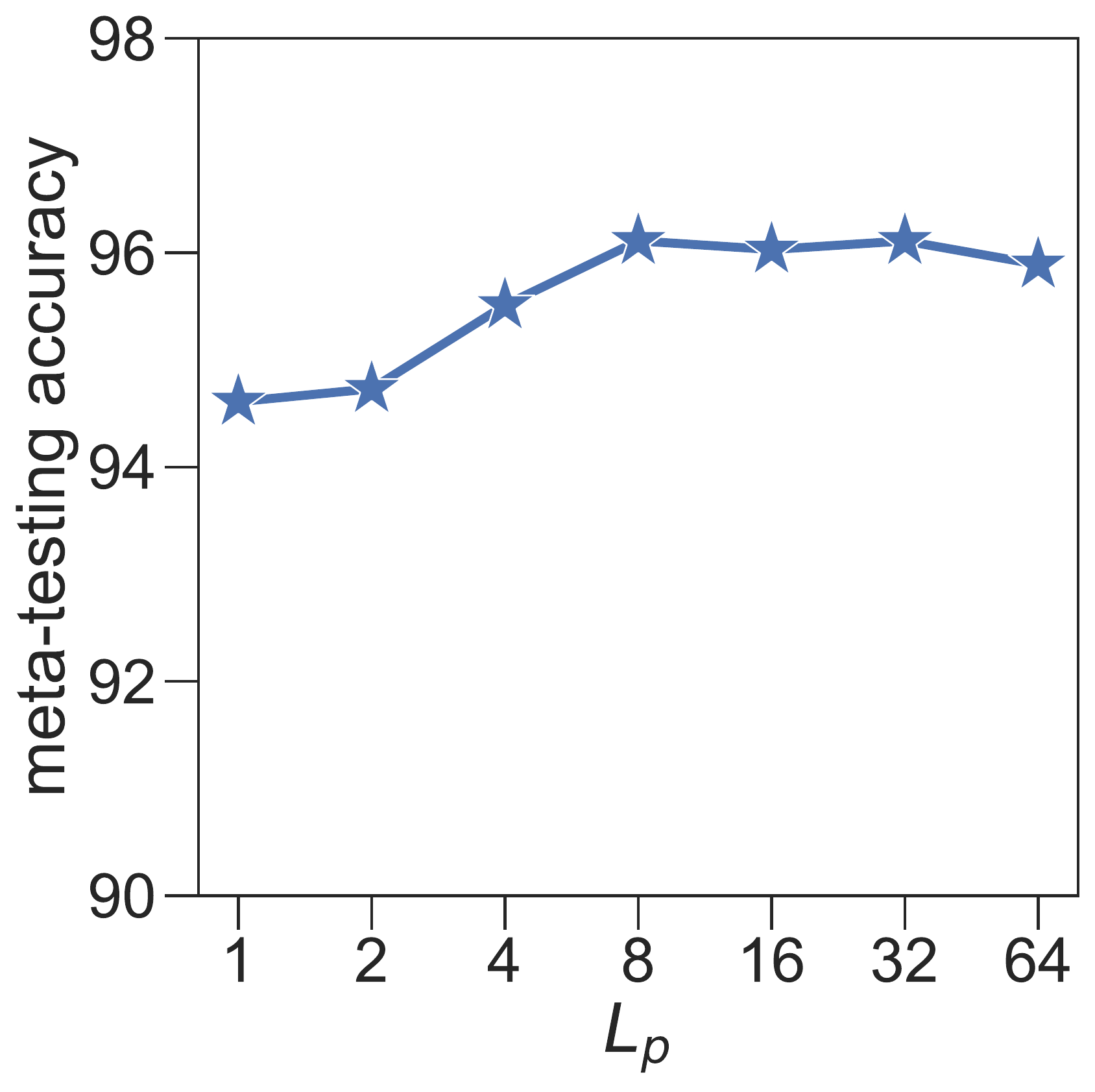}}\!\!
		\vskip -.15in
		\caption{Effect of $L_p$ (in log-scale) on 5-way 5-shot classification 
			($K=8$).}
		\label{fig:ab-Lp}
	\end{figure*}

	\begin{table*}[!h]
		\vskip -.05in
		\centering
		\caption{5-way 5-shot  classification meta-testing accuracy of MetaPrompter
		with different verbalizers.
		} \label{table:main-comp-verb}
		\resizebox{.8\textwidth}{!}{
			\begin{tabular}{cc| c c c c c c }
				\toprule
				\multicolumn{2}{c|}{verbalizer} & \multirow{2}{*}{\textit{20News} } &\multirow{2}{*}{\textit{Amazon} } & \multirow{2}{*}{\textit{HuffPost} }  &\multirow{2}{*}{\textit{Reuters} } & \multirow{2}{*}{\textit{HWU64} }&\multirow{2}{*}{\textit{Liu54} } \\
				hand-crafted & RepVerb & &&&&&\\
				\midrule
				\cmark & \xmark &85.91&81.96&70.37&95.91&91.89&90.32 \\ 
				\xmark & \cmark & 87.12 &	86.05&72.63&96.69&95.25& 93.35\\
				\cmark & \cmark& $\mathbf{88.57}$ & $\mathbf{86.36}$ & $\mathbf{74.89}$ & $\mathbf{97.63} $&$\mathbf{ 95.30}$& $\mathbf{95.47}$\\
				\bottomrule
			\end{tabular}
		}
	\end{table*}
	
	\begin{table*}[!h]
		\vskip -.05in
		\centering
		\caption{
			5-way 5-shot  classification meta-testing accuracy by using BMG to
	learn the prompt pool.}
		\label{table:bmg}
		\resizebox{.9\textwidth}{!}{
			\begin{tabular}{c c c c c c c c }
				\toprule
				&\#param $(\times 10^6)$ &  \textit{20News} &\textit{Amazon}& \textit{HuffPost}&  \textit{Reuters} & \textit{HWU64} & \textit{Liu54} \\
				\midrule
				MetaPrompting+BMG &$109.52$ & 	$85.71$ &$83.47$ &$73.92$ & $96.27$  & $93.31$ & 	$93.04$ \\
				MetaPrompter+BMG &$0.06$ & $\mathbf{87.91}$ & $\mathbf{86.45}$ & $\mathbf{74.99}$& $\mathbf{98.01}$  & $\mathbf{95.41}$ & $\mathbf{94.52}$ \\
				\bottomrule
			\end{tabular}
		}
	\end{table*}

	\subsection{Ablation Study}
	In this section,
	we perform ablation study
	using the 5-way 5-shot
	setting in Section \ref{sec:eval-MetaPrompter}.
	
	\subsubsection{Effect of $K$}
	Figure \ref{fig:ab-K}
	shows the 5-way 5-shot meta-testing accuracy 
	of MetaPrompter
	with varying $K$.
	As $K$ increases,
	more task knowledge can be extracted and 
	the meta-testing accuracy increases. However, using a 
	very large $K$ (e.g., $64$) is unnecessary and the
	accuracy flattens.
	
	\subsubsection{Effect of $L_p$}
	Figure \ref{fig:ab-Lp}
	shows
	the 5-way 5-shot meta-testing accuracy 
	of MetaPrompter
	with varying $L_p$.
	As $L_p$ increases, the meta-testing accuracy
	increases  
	as
	the expressive power of the prompt pool is enhanced. 
	However, using a very large
	$L_p$  is 
	again unnecessary 
	and the accuracy flattens.
	
	\subsubsection{Effect of Verbalizer}
	Table \ref{table:main-comp-verb}
	shows the 
	number of parameters 
	and 
	meta-testing accuracy of MetaPrompter with
	hand-crafted verbalizer
	(used in
	\eqref{eq:total-pred})
	and RepVerb.
	As can be seen, RepVerb
	is better than the hand-crafted
	verbalizer, and
	combining both yields the best result.
	
	\subsubsection{Integration with Other Meta-learning Algorithms}
	While the MAML algorithm~\citep{Finn2017} is used in Algorithm \ref{alg},
	other meta-learning algorithms 
	can also be used 
	to learn the prompt pool 
	in MetaPrompter 
	or
	the meta-initialized prompt
	in MetaPrompting.
	In this experiment, we replace MAML with the state-of-the-art
	BMG 
	\citep{Flennerhag2022}.
	Table \ref{table:bmg} shows the meta-testing accuracy and number of parameters.
	As can be seen, MetaPrompter+BMG consistently outperforms MetaPrompting+BMG.

	%%%%%%%%%%%%%%%%%%%%%%%%%%%%%
	%   section: conclusion
	%%%%%%%%%%%%%%%%%%%%%%%%%%%%%
	\section{Conclusion }
	In this paper,
	we proposed
	MetaPrompter, an effective and parameter-efficient algorithm
	for prompt tuning.
	It combines structured prompting and a novel 
	verbalizer called RepVerb.
	A prompt pool structure is used to construct instance-dependent prompts
	by attention,
	while
	RepVerb
	builds label embedding
	by averaging feature embeddings of the corresponding
	training samples.
	The pool of prompts  is meta-learned from the meta-training tasks.
	Experimental results 
	demonstrate the effectiveness of 
	the proposed
	MetaPrompter
	and RepVerb.	
	
	One limitation is that MetaPrompter is based on meta-learning, and so requires the availability of a set of
	meta-training tasks.

	\section*{Acknowledgements}
	This work was supported by NSFC key grant 62136005, NSFC general grant 62076118, and Shenzhen fundamental research program JCYJ20210324105000003.
	This research was supported in part by the Research Grants Council of the Hong Kong Special Administrative Region (Grant 16200021).

	\bibliography{paper}
	\bibliographystyle{icml2023}

	%%%%%%%%%%%%%%%%%%%%%%%%%%%%%%%%%%%%%%%%%%%%%%%%%%%%%%%%%%%%%%%%%%%%%%%%%%%%%%%
	%%%%%%%%%%%%%%%%%%%%%%%%%%%%%%%%%%%%%%%%%%%%%%%%%%%%%%%%%%%%%%%%%%%%%%%%%%%%%%%
	% APPENDIX
	%%%%%%%%%%%%%%%%%%%%%%%%%%%%%%%%%%%%%%%%%%%%%%%%%%%%%%%%%%%%%%%%%%%%%%%%%%%%%%%
	%%%%%%%%%%%%%%%%%%%%%%%%%%%%%%%%%%%%%%%%%%%%%%%%%%%%%%%%%%%%%%%%%%%%%%%%%%%%%%%
	\newpage
	\appendix
	\onecolumn

	\section{Implementation of WARP}
	\label{sec:app-warp}
	WARP~\cite{hambardzumyan2021warp}
	is developed for supervised learning with limited samples. 
	The proposed RepVerb 
	(Algorithm \ref{alg:repverb})
	is also designed for the supervised learning setting.
	In the
	meta-learning procedure in
	Algorithm \ref{alg},
	it is used in 
	the inner level 
	(steps \ref{step:init}-\ref{alg:base-b}) which
	is also supervised.
	
	Given a meta-testing
	task $\tau'=(\hS_{\tau'}, \hQ_{\tau'})$ with label set $\hY_{\tau'}$,
	let $\vV\equiv\{\vv_y: y \in \hY_{\tau'}\}$ be $\tau'$'s learnable label
	embeddings, and $\vphi$ be the
	MLM
	parameter.
	For an input $\vx$,
	the 
	distribution 
	for labels $y\in \hY_{\tau'}$
	is predicted as:
	\begin{align}
		\bP(y|\vx; \vphi, \vV) = \frac{\exp(\vv_y^\top\vh_{\tttext{[MASK]}}(\tilde{\vx}))}{\sum_{y'\in \hY_\tau}\exp(\vv_{y'}^\top\vh_{\tttext{[MASK]}}(\tilde{\vx}))}, 
		\label{eq:warp}
	\end{align}
	where $\vh_{\tttext{[MASK]}}(\tilde{\vx})$ is the $\tttext{[MASK]}$'s embedding of wrapped input $\tilde{\vx}$.
	$(\vphi, \vV)$ 
	is learned  by performing 
	$T=5$ gradient updates
	to minimize the negative log-likelihood loss
	on the support set $\hS_{\tau'}$:
	\begin{align*}
		\sum_{(\vx,y) \in \hS_{\tau'}} - \log \bP(y|\vx; \vphi, \vV).
	\end{align*}
	$\vphi$ is initialized by the pre-trained MLM,
	while $\vV$ is initialized randomly.
	The learned $(\vphi, \vV)$ is then evaluated on $\hQ_{\tau'}$.
	For a test sample $(\vx^\star, \cdot) \in \hQ_{\tau'}$, its prediction is
	given by \eqref{eq:warp}.	
	We run
	the WARP algorithm 
	on all meta-testing tasks and report the average meta-testing accuracy in Table \ref{table:verb}.
	
	\section{Visualization  for Verbalizers}
	\label{sec:app-verb}

	\begin{figure*}[!h]
		\centering
		\vskip -.2in
		\subfigure[WARP on \textit{20News}.\label{fig:warp-News20}]{\includegraphics[width=0.19\textwidth]{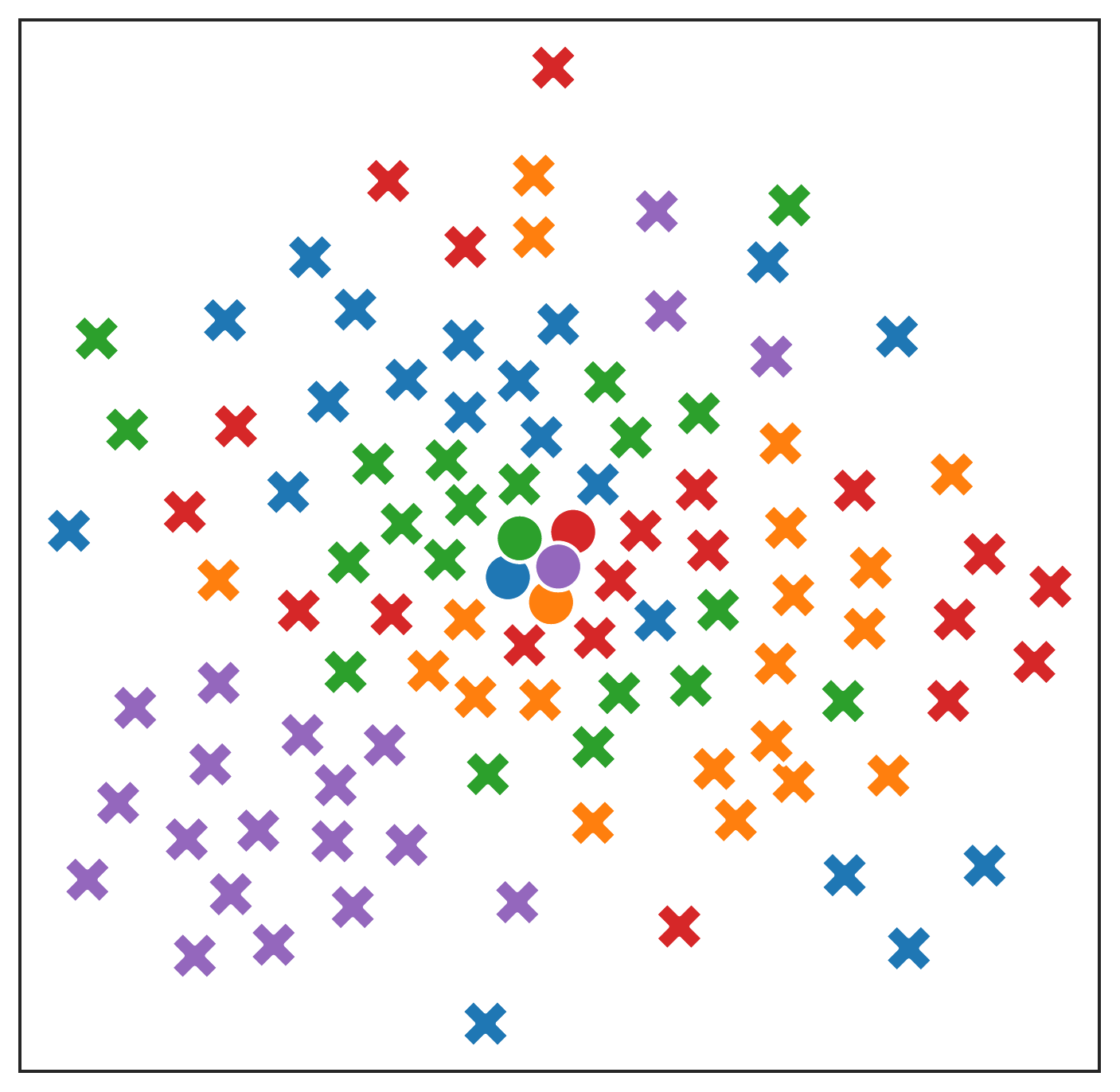}}
		\subfigure[ProtoVerb on \textit{20News}.\label{fig:protoverb-News20}]{\includegraphics[width=0.19\textwidth]{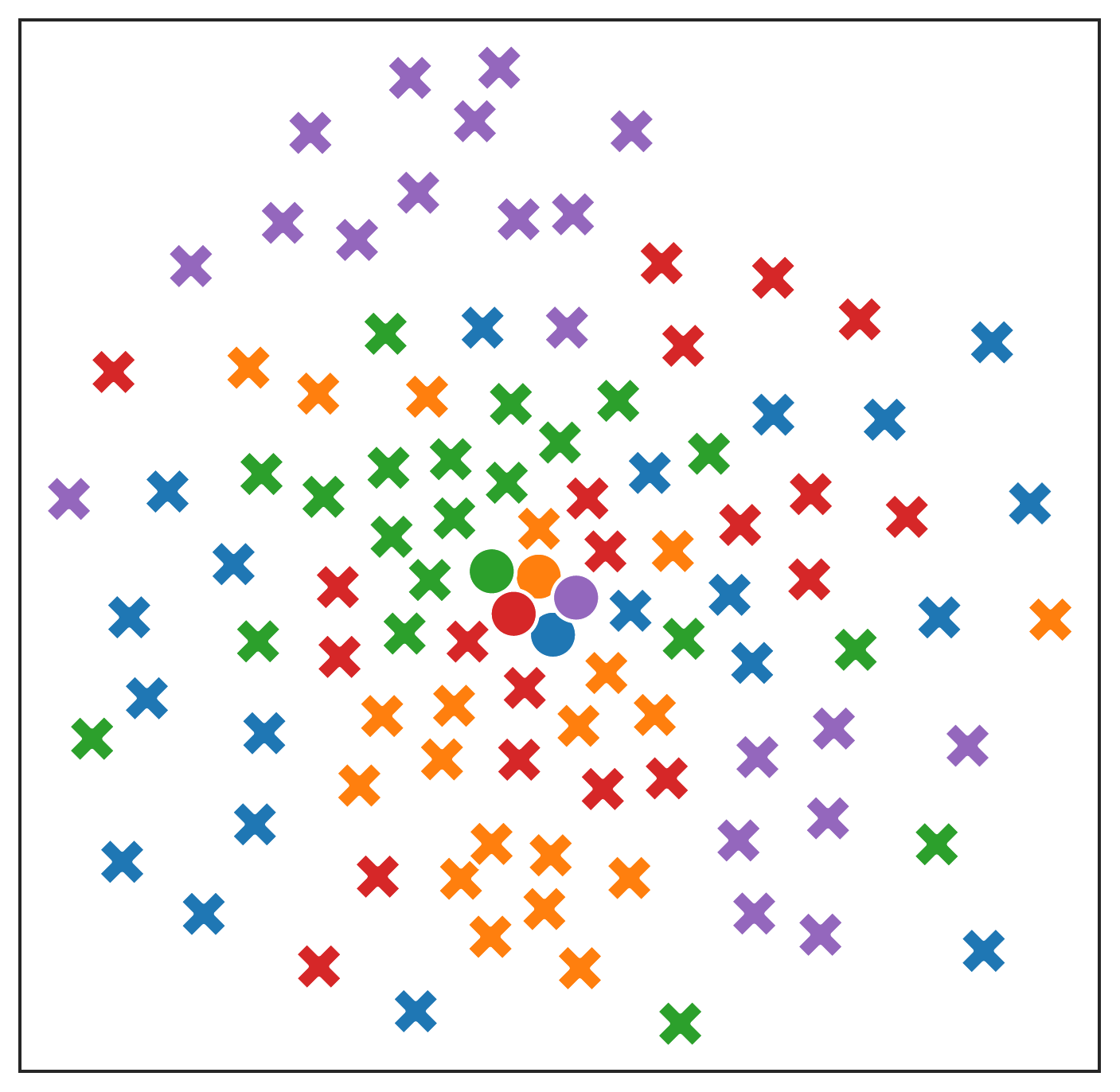}}
		\subfigure[RepVerb on \textit{20News}.\label{fig:repverb-News20}]{\includegraphics[width=0.19\textwidth]{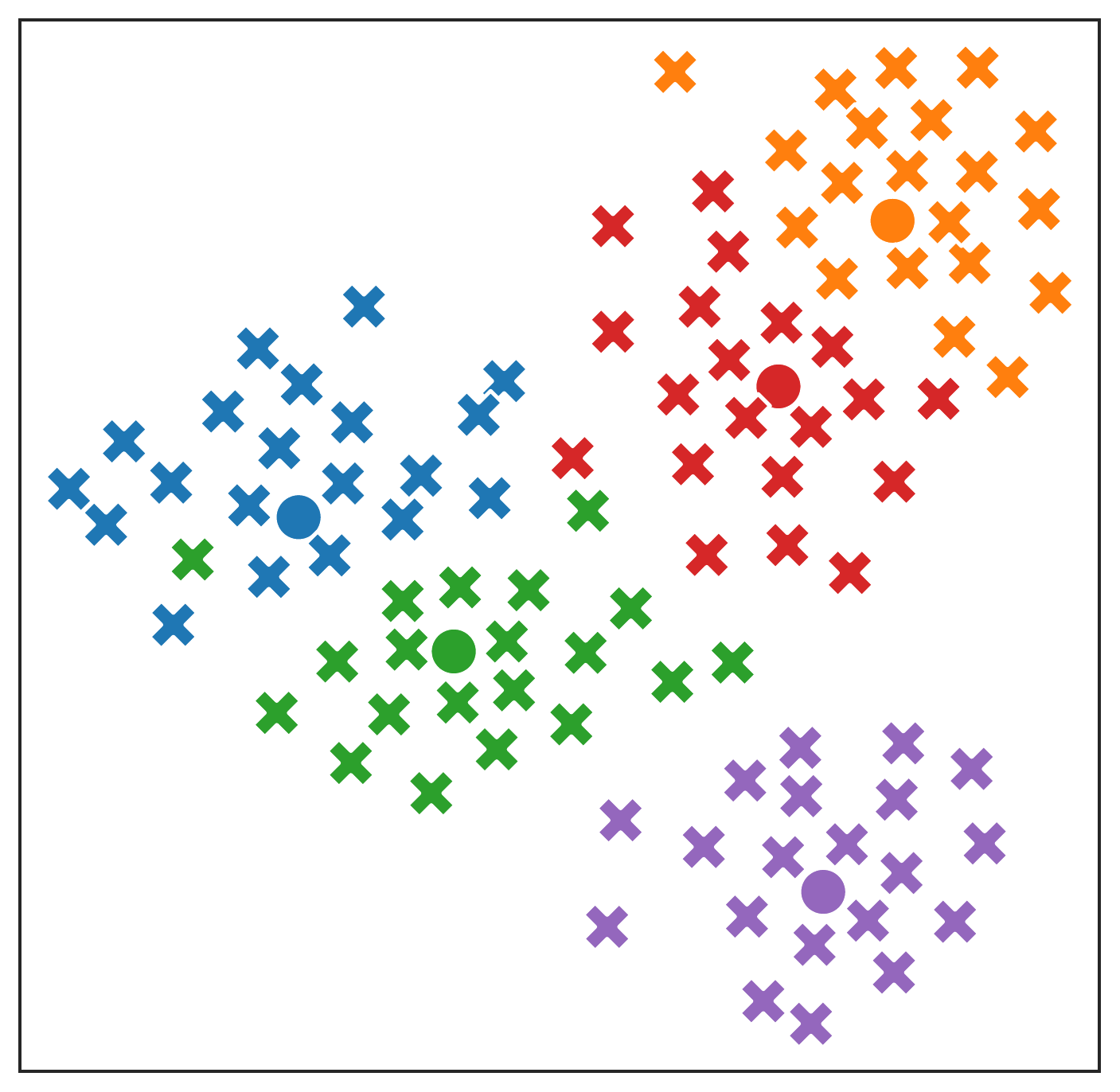}} \\
		\vskip -.1in
		\subfigure[WARP on \textit{Amazon}.\label{fig:warp-Amazon}]{\includegraphics[width=0.19\textwidth]{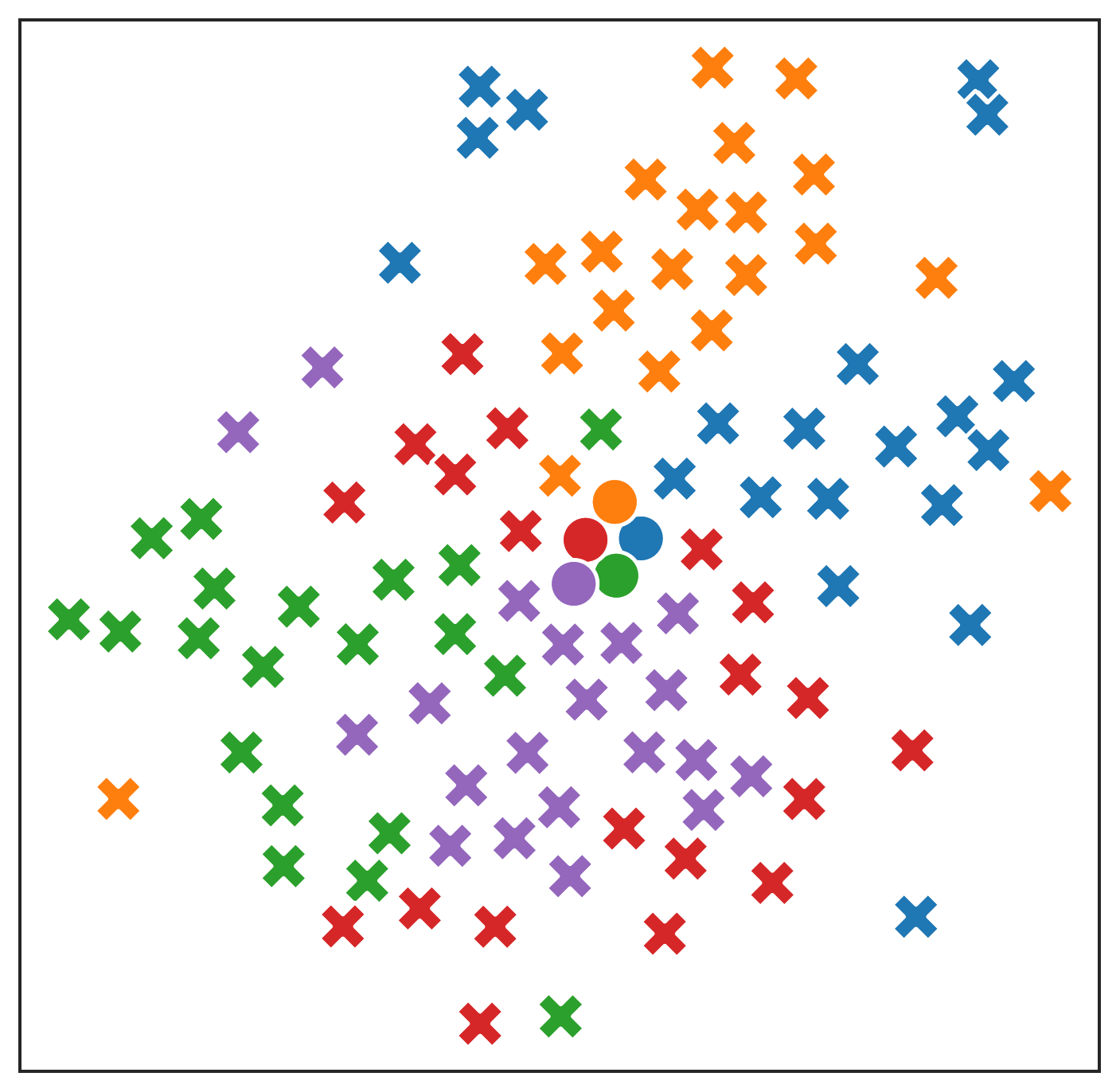}}
		\subfigure[ProtoVerb on \textit{Amazon}.\label{fig:protoverb-Amazon}]{\includegraphics[width=0.19\textwidth]{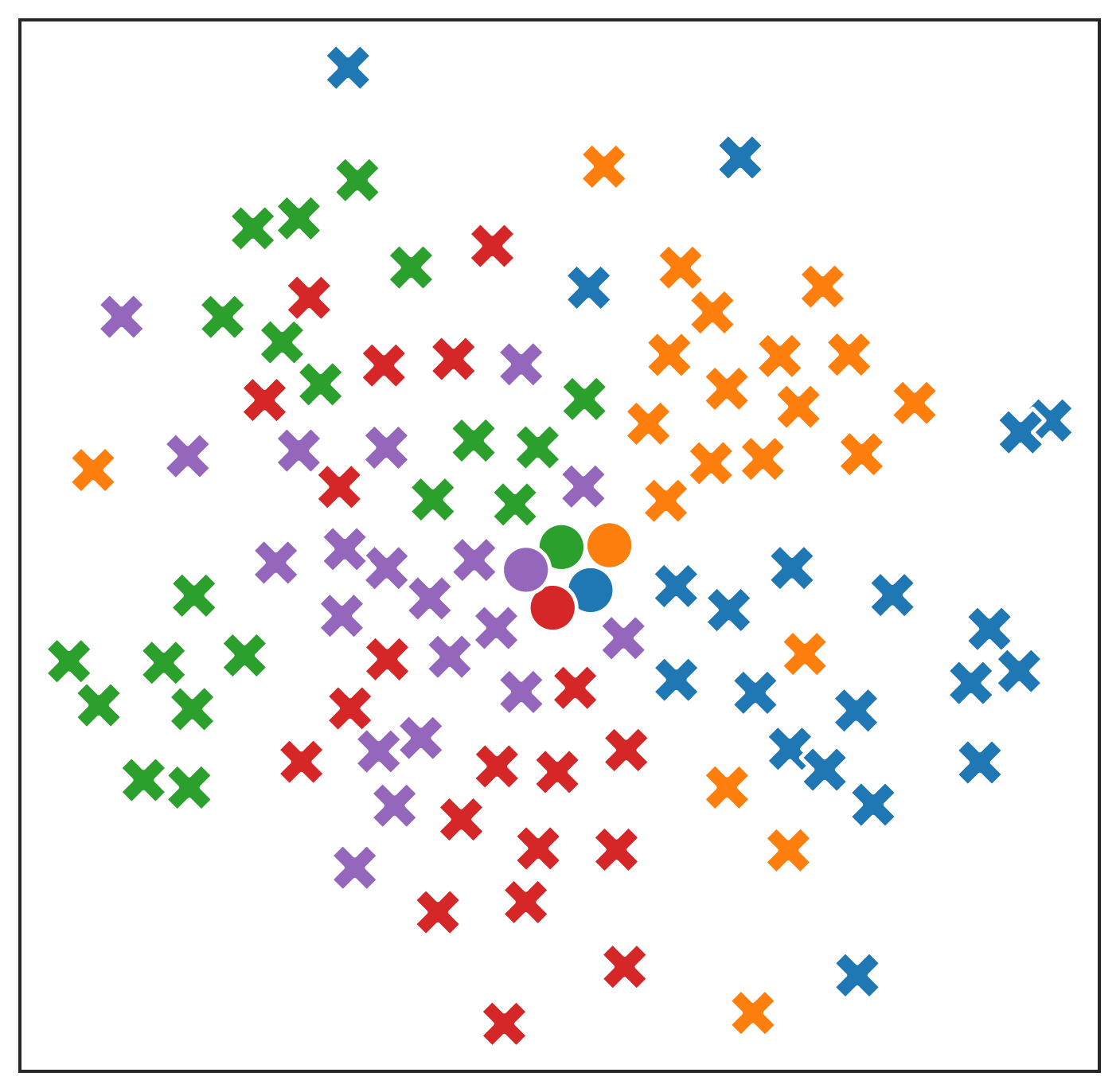}}
		\subfigure[RepVerb on \textit{Amazon}.\label{fig:repverb-Amazon}]{\includegraphics[width=0.19\textwidth]{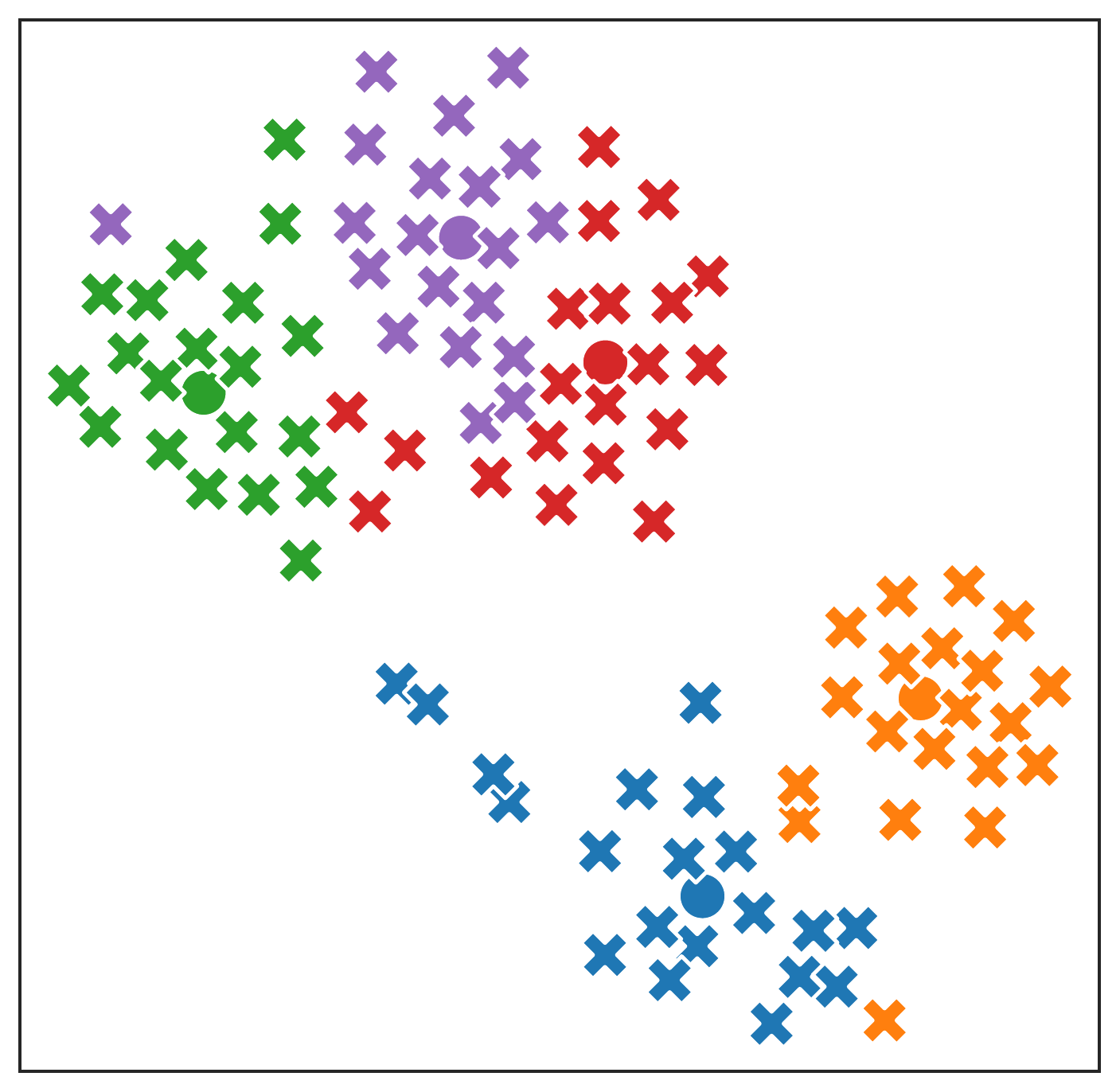}} \\
		\vskip -.1in
		\subfigure[WARP on \textit{HuffPost}.\label{fig:warp-HuffPost}]{\includegraphics[width=0.19\textwidth]{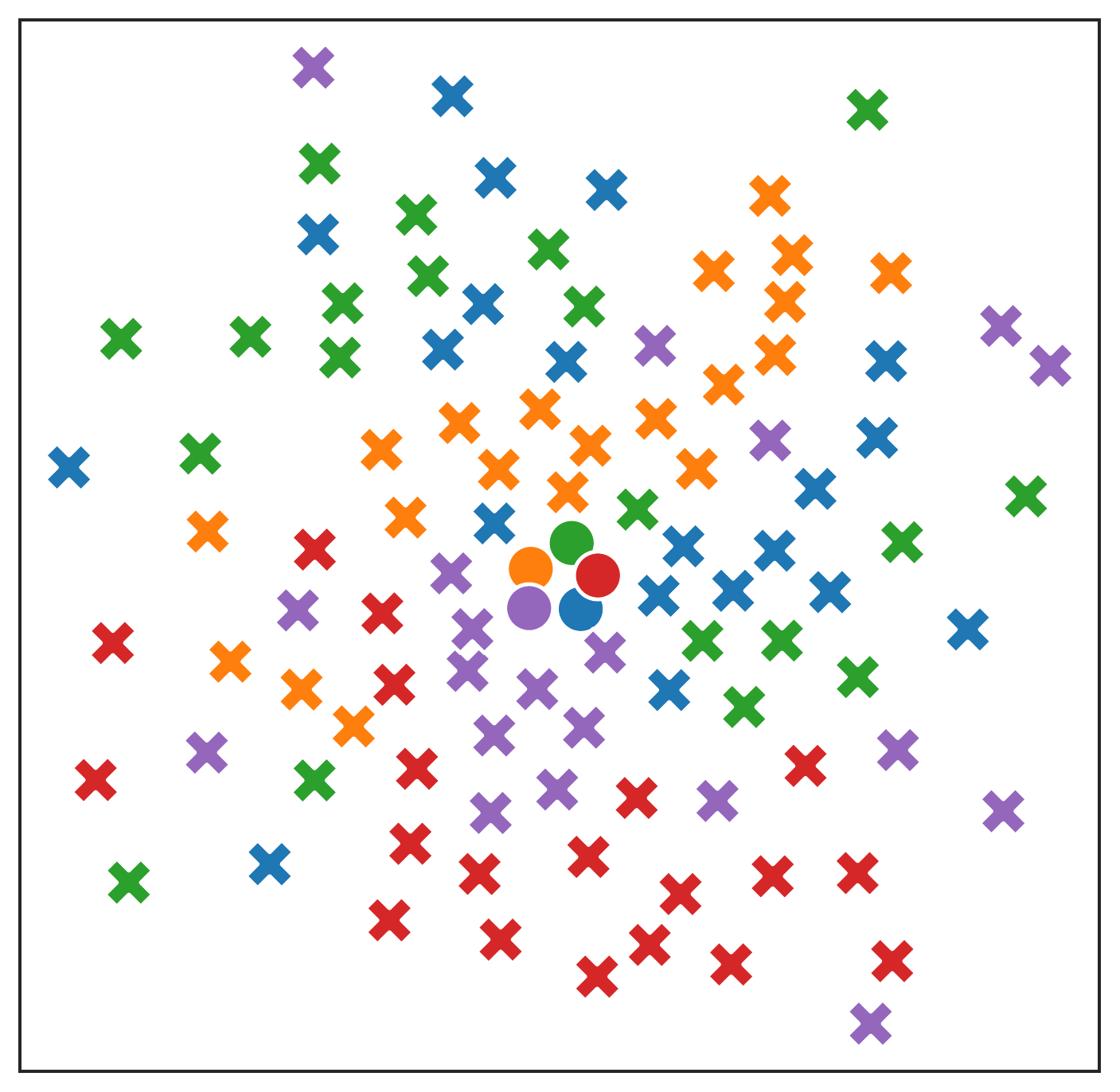}}
		\subfigure[\!\!ProtoVerb on \textit{HuffPost}.\!\label{fig:protoverb-HuffPost}]{\includegraphics[width=0.19\textwidth]{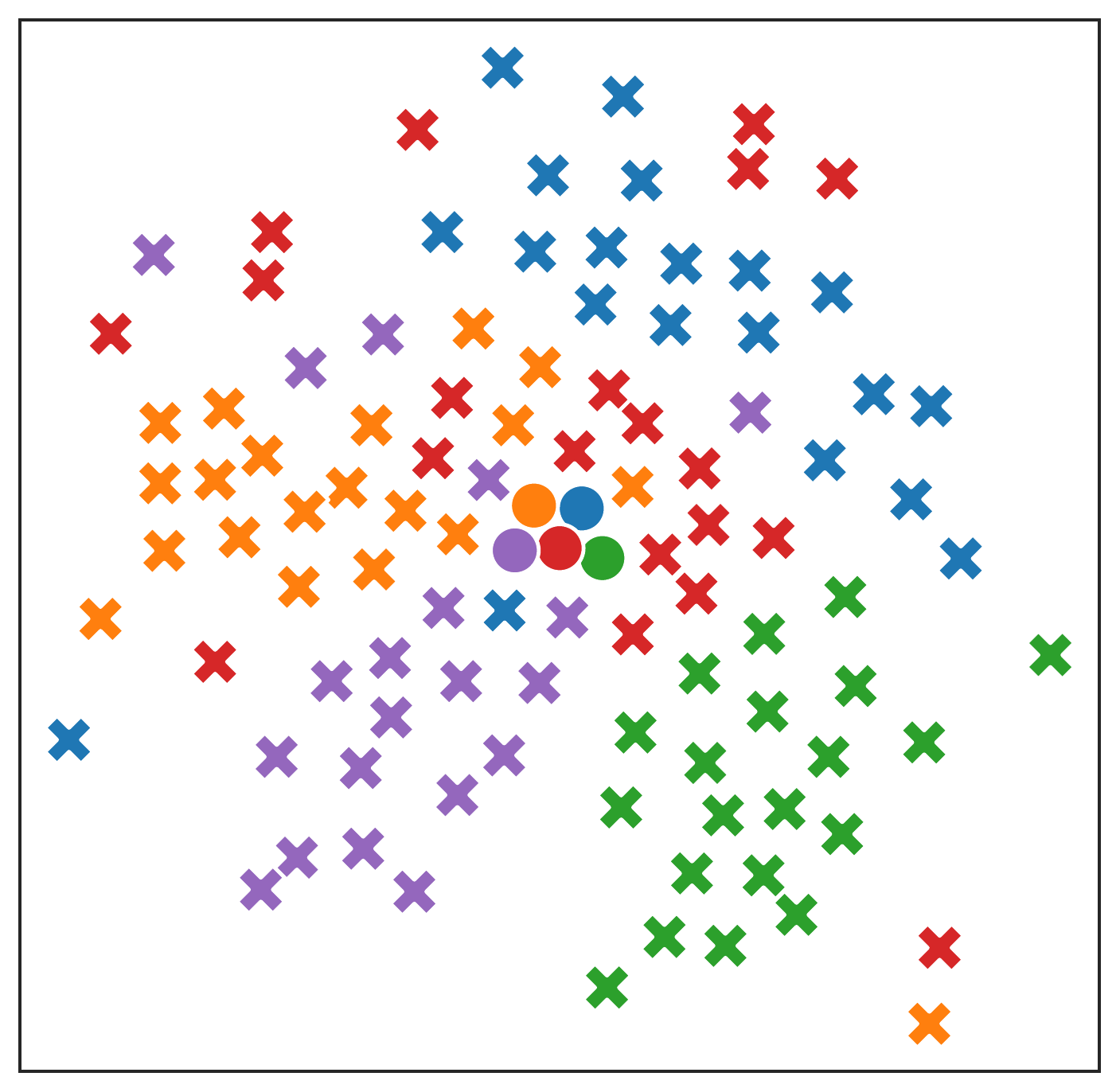}}
		\subfigure[RepVerb on \textit{HuffPost}.\label{fig:repverb-HuffPost}]{\includegraphics[width=0.19\textwidth]{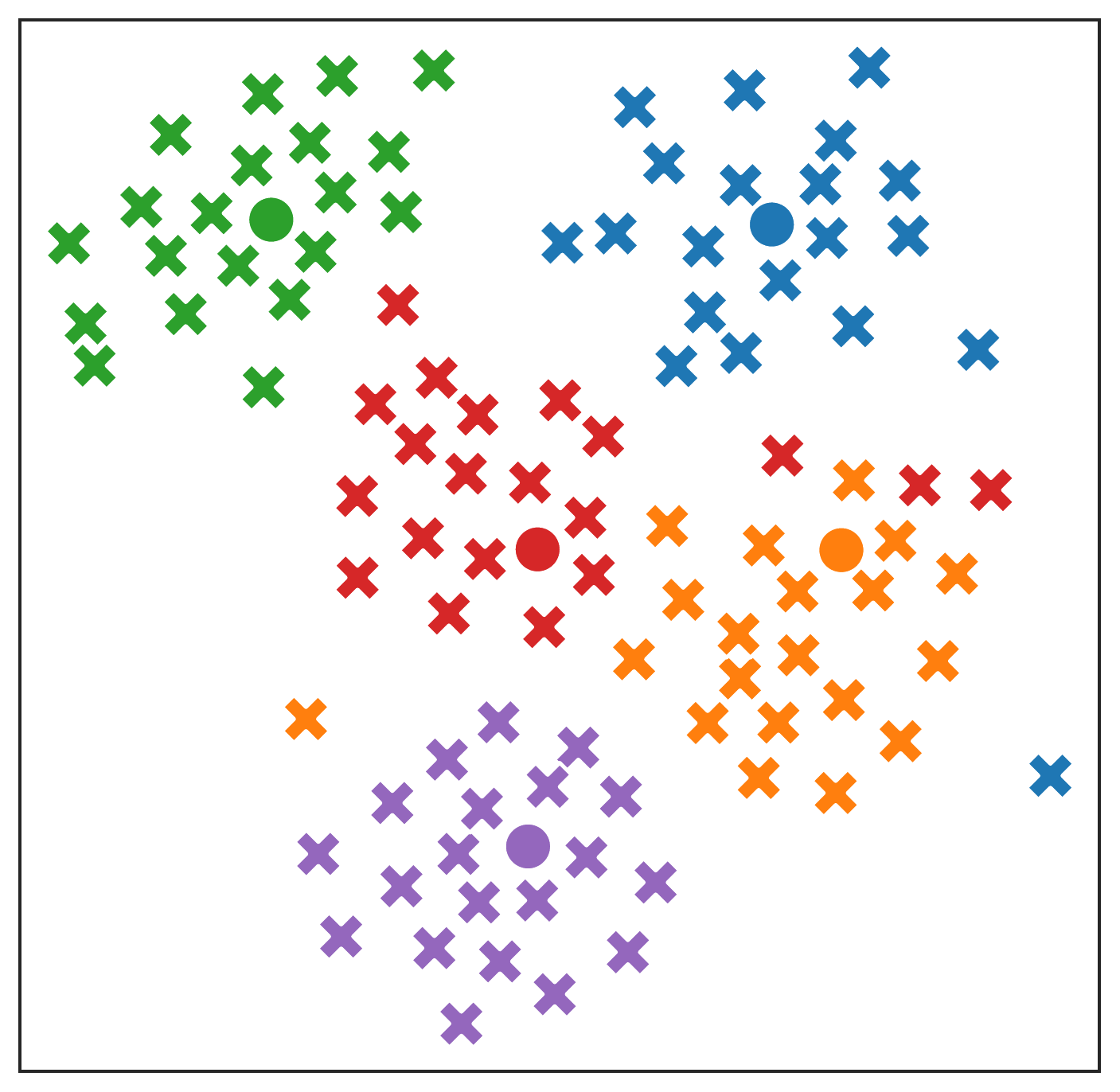}} \\
		\vskip -.1in
		\subfigure[WARP on \textit{HWU64}.\label{fig:warp-hwu}]{\includegraphics[width=0.19\textwidth]{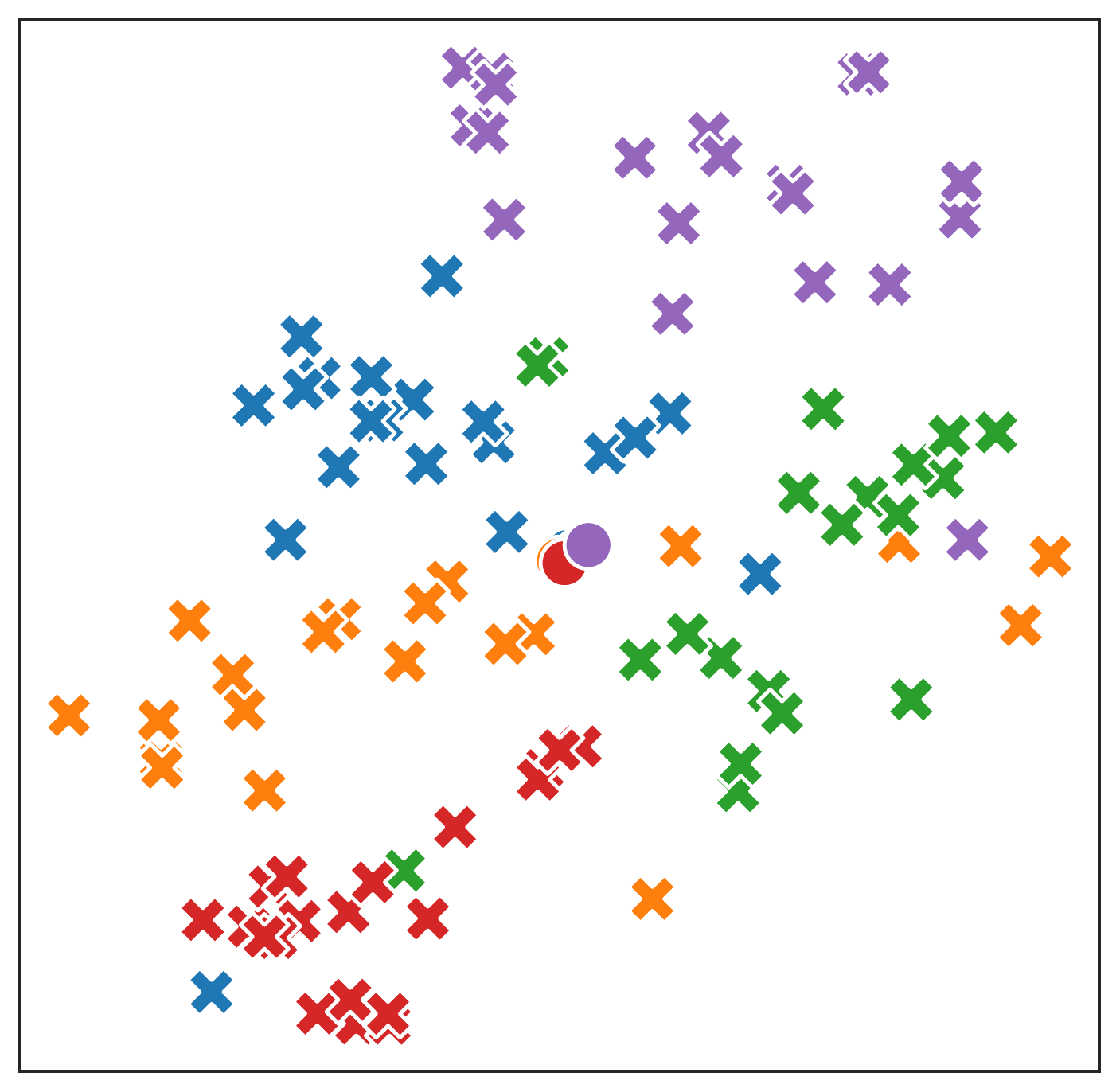}}
		\subfigure[ProtoVerb on \textit{HWU64}.\label{fig:protoverb-hwu}]{\includegraphics[width=0.19\textwidth]{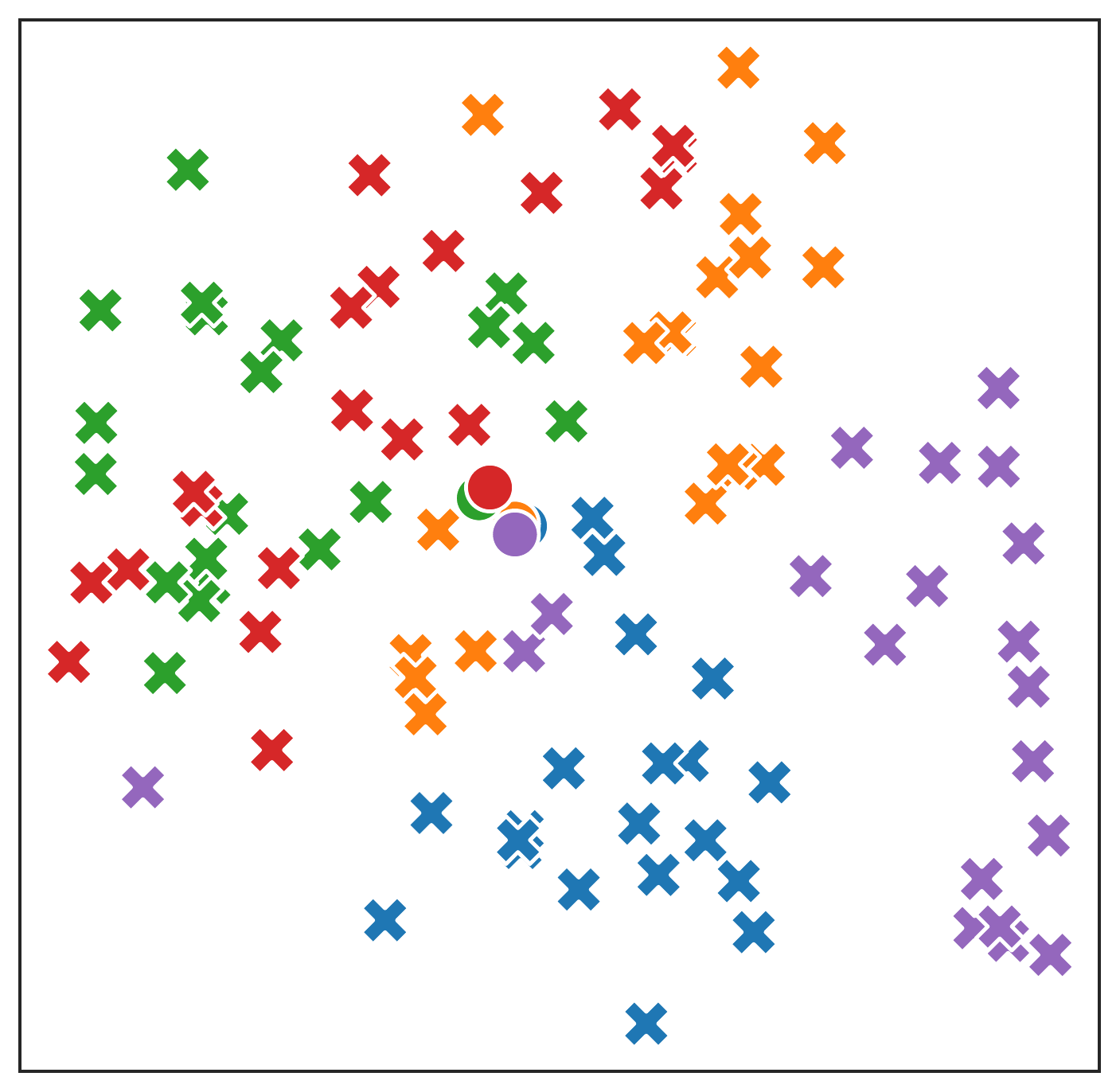}}
		\subfigure[RepVerb on \textit{HWU64}.\label{fig:repverb-hwu}]{\includegraphics[width=0.19\textwidth]{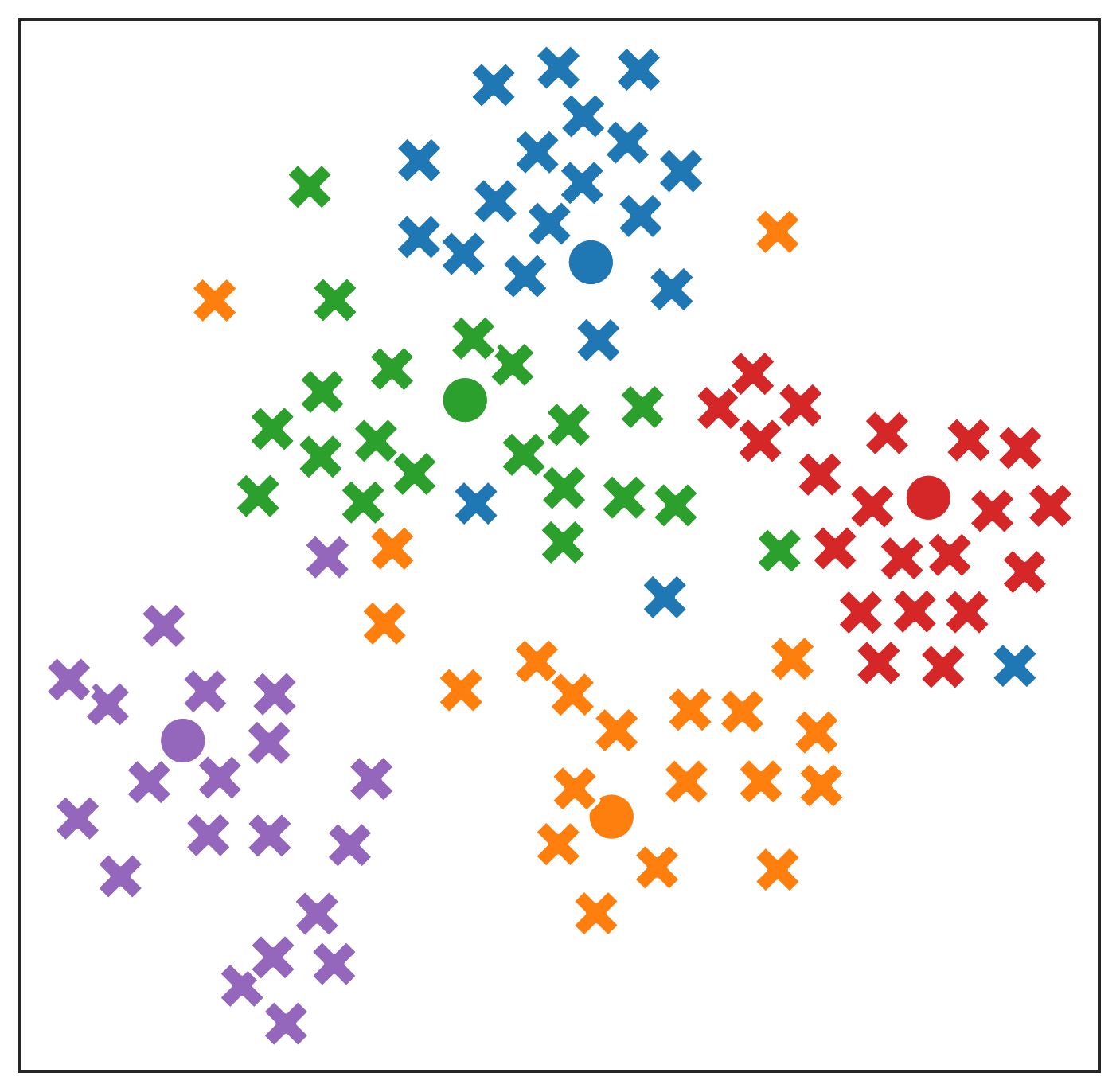}} \\
		\vskip -.1in
		\subfigure[WARP on \textit{Liu54}.\label{fig:warp-liu}]{\includegraphics[width=0.19\textwidth]{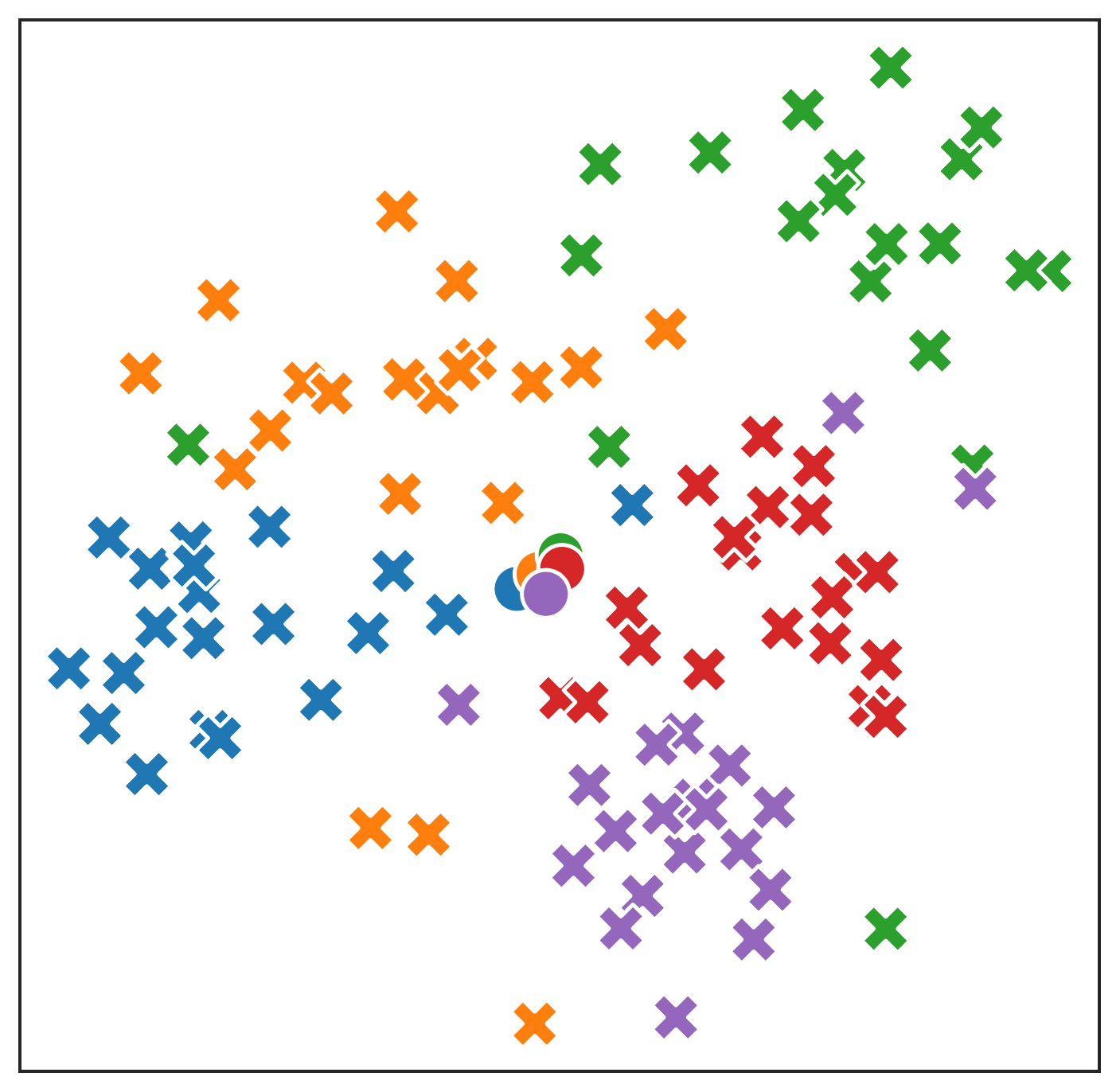}}
		\subfigure[ProtoVerb on \textit{Liu54}.\label{fig:protoverb-liu}]{\includegraphics[width=0.19\textwidth]{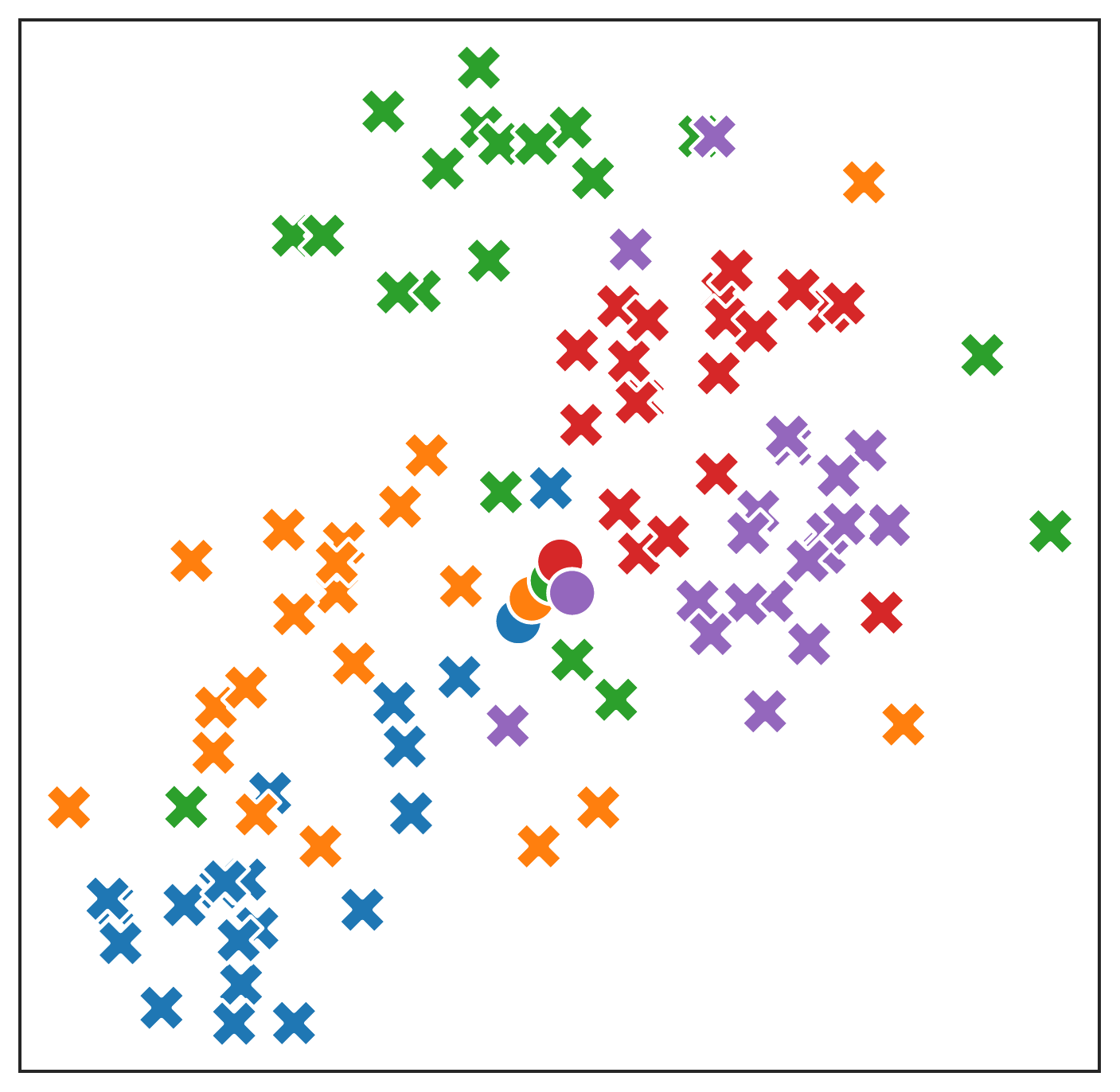}}
		\subfigure[RepVerb on \textit{Liu54}.\label{fig:repverb-liu}]{\includegraphics[width=0.19\textwidth]{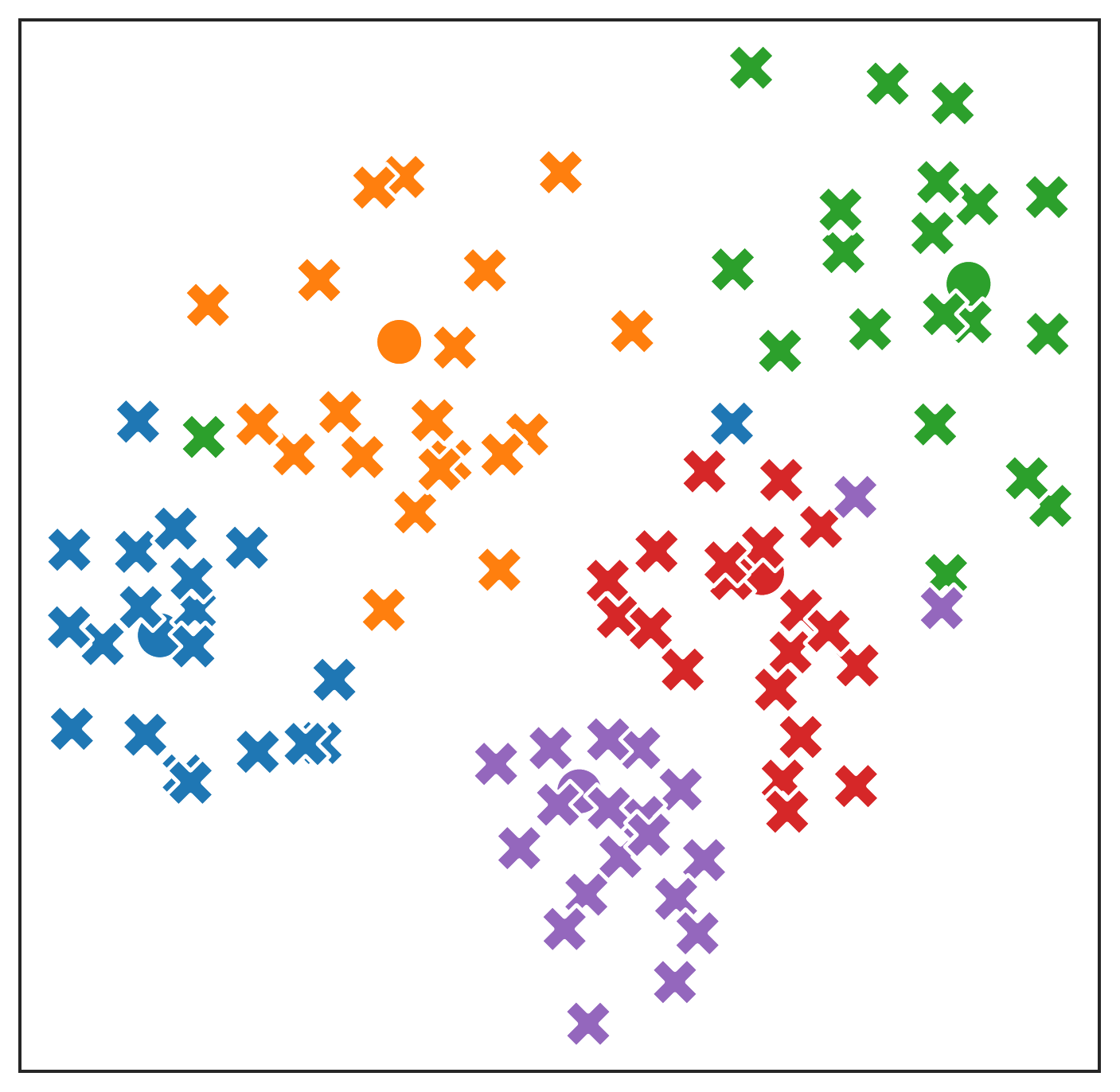}}
		\vskip -.1in
		\caption{t-SNE visualization of \tttext{[MASK]}'s embeddings (crosses) and
			label embeddings (circles) for a 5-way 5-shot task randomly sampled from \textit{20News}, \textit{Amazon}, \textit{HuffPost}, \textit{HWU64}, and \textit{Liu54}.}
		\label{fig:visualize-others}
	\end{figure*}
	
	Figure \ref{fig:visualize-others}
	shows the
	t-SNE
	visualization of the embeddings ($\vh_{\tttext{[MASK]}}(\vx)$'s)
	of
	$100$ samples ($\vx$'s)
	and learned
	label embeddings
	($\vv_y$'s)
	of a 5-way 5-shot 
	task randomly from \textit{20News}, \textit{Amazon}, \textit{HuffPost}, \textit{HWU64}, and \textit{Liu54}.
	As shown, the
	RepVerb embedding is more discriminative and compact
	than WARP and ProtoVerb.
	Furthermore, RepVerb's label embedding is consistent with the samples'
	feature embeddings, while those of
	WARP and ProtoVerb are not.

\end{document}